\documentclass{article}
\usepackage[nonatbib, preprint]{arxiv}

\usepackage{grffile}
\usepackage[T1]{fontenc}
\usepackage[utf8]{inputenc}
\usepackage{graphicx}
\usepackage[english]{babel}
\usepackage{booktabs}
\usepackage{amsmath,amssymb,amsfonts}
\usepackage{amsthm}
\usepackage{algorithmic}
\usepackage{textcomp}
\usepackage{hyperref}
\usepackage{diagbox}
\usepackage{color}
\usepackage{xcolor}
\usepackage{colortbl}
\usepackage[flushleft]{threeparttable}
\usepackage{amsmath,amsfonts,amssymb}
\usepackage{tikz}
\usepackage{xspace}
\usepackage{bbding} 
\usepackage{romannum}
\usepackage{enumitem}
\usepackage{caption}
\usepackage{subcaption}
\usepackage[ruled,linesnumbered,noend]{algorithm2e}
\usepackage{graphicx}
\hypersetup{
    colorlinks=true,
    linkcolor=blue,
    filecolor=magenta,      
    urlcolor=cyan,
    pdftitle={Overleaf Example},
    pdfpagemode=FullScreen,
    }

\begin{document}
\title{Fast and Attributed Change Detection on Dynamic Graphs with Density of States}


\author{%
  Shenyang Huang\\
  McGill University, Mila\\
  \texttt{shenyang.huang@mail.mcgill.ca} \\
  \And
  Jacob Danovitch\\
  McGill University, Mila\\
  \texttt{jacob.danovitch@mail.mcgill.ca} \\
  \And
  Guillaume Rabusseau\\
  DIRO, Universit\'e de Montr\'eal, Mila \\
  CIFAR AI chair \\
  \texttt{guillaume.rabusseau@umontreal.ca} \\
  \And
  Reihaneh Rabbany\\
  McGill University, Mila \\
  CIFAR AI chair \\
  \texttt{reihaneh.rabbany@mila.quebec} \\
}
\maketitle              

\SetKwInOut{Parameter}{parameter}
\SetKwIF{If}{ElseIf}{Else}{if}{}{else if}{else}{end if}%
\SetKwFor{While}{while}{}{end while}%
\SetKwRepeat{Do}{do}{while}

\newcommand\thefont{\expandafter\string\the\font}


\newcommand*\circled[1]{\tikz[baseline=(char.base)]{
            \node[shape=circle,draw,inner sep=1pt] (char) {#1};}}

\newcommand{\method}{LAD\xspace}

\newcommand{\newmethod}{SCPD\xspace}


\newcommand{\etc}{et al.\xspace}
\renewcommand{\vec}[1]{\ensuremath{\mathbf{#1}}}
\newcommand{\vecs}[1]{\ensuremath{\mathbf{\boldsymbol{#1}}}}
\newcommand{\mat}[1]{\ensuremath{\mathbf{#1}}}
\newcommand{\mats}[1]{\ensuremath{\mathbf{\boldsymbol{#1}}}}
\newcommand{\ten}[1]{\mat{\ensuremath{\boldsymbol{\mathcal{#1}}}}}

\newcommand{\vectorize}[1]{\mathrm{vec}\left(#1\right)}
\newcommand{\rank}[1]{\mathrm{rank}(#1)}
%
\newcommand{\kron}{\otimes}
\newcommand{\krao}{\odot}
\newcommand{\outprod}{\circ}
\newcommand{\mlrank}[1]{\mathrm{rank}_{\textit{ml}}(#1)}
\newcommand{\cprank}[1]{\mathrm{rank}_{\textit{CP}}(#1)}
\newcommand{\ttm}[1]{\times_{#1}}
\newcommand{\ttv}[1]{\bullet_{#1}}
\newcommand{\tenmat}[2]{#1_{(#2)}}
\newcommand{\CP}[1]{\llbracket #1 \rrbracket}
\newcommand{\Tucker}[1]{\llbracket #1 \rrbracket}
\newcommand{\TT}[1]{\llangle #1 \rrangle}
\newtheorem{definition}{Definition}

\newcommand{\txthead}[1]{\underline{\emph{#1}}}

\newcommand{\changed}[1]{\textcolor{black}{#1}}
\pagenumbering{arabic} 
\begin{abstract}

\changed{How can we detect traffic disturbances from international flight transportation logs or changes to collaboration dynamics in academic networks? These problems can be formulated as detecting anomalous change points in a dynamic graph. Current solutions do not scale well to large real-world graphs, lack robustness to large amounts of node additions/deletions, and overlook changes in node attributes. To address these limitations, we propose a novel spectral method: Scalable Change Point Detection~(\newmethod). \newmethod generates an embedding for each graph snapshot by efficiently approximating the distribution of the Laplacian spectrum at each step. \newmethod can also capture shifts in node attributes  by tracking correlations between attributes and eigenvectors. Through extensive experiments using synthetic and real-world data, we show that \newmethod (a)  achieves state-of-the-art performance, (b) is significantly faster than the state-of-the-art methods and can easily process millions of edges in a few CPU minutes, (c)  can effectively tackle a large quantity of node attributes, additions or deletions and (d)  discovers interesting events in large real-world graphs. The code is publicly available at \url{https://github.com/shenyangHuang/SCPD.git}. }
\end{abstract}

\section{Introduction}\label{Sec:intro}


Anomaly detection is one of the fundamental tasks in analyzing dynamic graphs~\cite{ranshous2016scalable,peel2015detecting,eswaran2018sedanspot}, with applications ranging from detecting disruptions in traffic networks, analyzing shifts in political environments, and identifying abnormal events in communication networks. In this work, we focus on identifying anomalous time points where the graph structure deviates significantly from the normal behavior, also known as change point detection~\cite{huang2020laplacian,ide2004eigenspace}. Detecting anomalies in dynamic graphs offers several challenges: real-world graphs are often very large, their size can drastically evolve over time (e.g., nodes appearing and disappearing in social network graphs where nodes represent users), and complex information is associated with nodes in the graph (e.g., profile of users summarized as a set of attributes for each node).

\begin{figure}[t]
    \begin{center}
        \centerline{\includegraphics[width=0.7\textwidth]{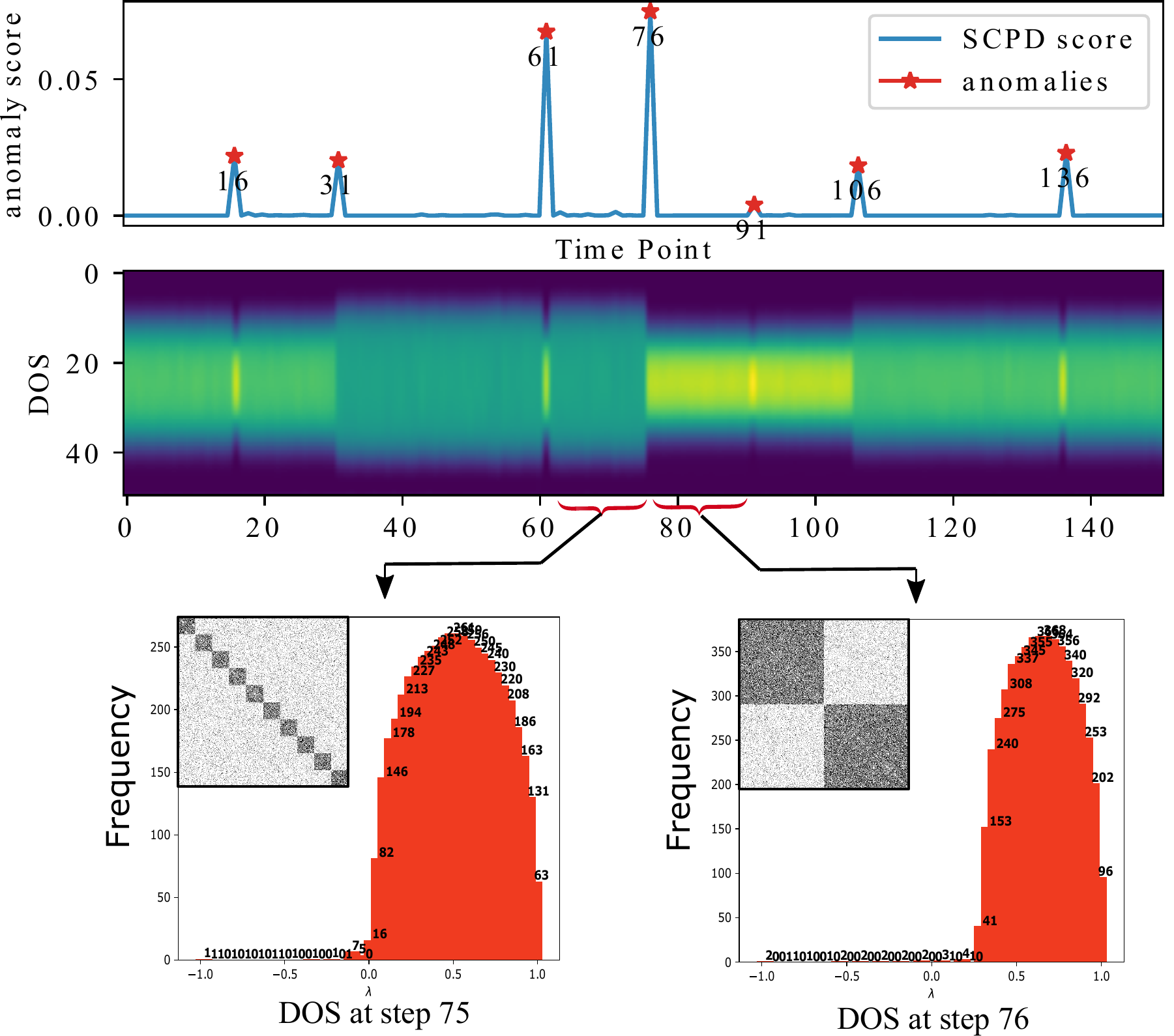}}
        \caption{ \changed{\newmethod utilizes the spectral density (approximated by Density of States~(DOS)) to summarize the graph at each time point, a change in DOS often indicates a change in graph distribution. The DOS becomes skewed after the number of communities decreases from ten to two in the SBM~\cite{holland1983stochastic} hybrid experiment~(see Section~\ref{Sec:synthetic}). The DOS is plotted for steps 75 and 76, while the inset plots show the adjacency matrix of the graph.}}
        \label{Fig:DOS.approx}
    \end{center}
    \vskip -0.5in
\end{figure}

Prior work on change point detection is limited by one or more of the following issues. \emph{1). Lack of scalability}: modern networks often contain millions of edges and nodes. Thus, computationally intensive algorithms~\cite{huang2020laplacian,koutra2012tensorsplat} can be challenging to apply on graphs with more than hundreds of nodes. \emph{2). Overlooking attributes}: many networks also contain a diverse set of node attributes which  evolve over time. No prior work has considered the evolution of node attributes and its relation with the graph structure. \emph{3). Difficulty with evolving sizes}: real networks grow over time, with new nodes often forming a large portion of the network. Methods such as~\cite{eswaran2018spotlight, wang2017fast} track a fixed set of nodes sampled from the initial time step and are thus limited to detecting changes happening within the initial set of nodes. Other approaches, such as~\cite{huang2020laplacian,ide2004eigenspace}, summarize each snapshot with a vector dependent on the size of the snapshot. Therefore, as the graph grows, truncation on the summary vector is required to ensure a uniform vector size for all snapshots.

To address the above limitations, we propose Scalable Change Point Detection~(\newmethod), a novel change point detection method which detects both structural and node attribute anomalies in  dynamic graphs. \newmethod utilizes the distribution of eigenvalues~(also known as the spectral density) of the Laplacian matrix as a low-dimensional embedding of each graph snapshot. As change points induce a shift in graph distribution, they also cause changes in the spectral density. We leverage the Density of States~(DOS)~\cite{dong2019network} framework to efficiently approximate the spectral density, allowing \newmethod to scale to dynamic graphs with millions of nodes. Figure~\ref{Fig:DOS.approx} illustrates the key idea of \newmethod: to discretize the spectral density, the range of eigenvalues is divided into $k$ bins, and the number of eigenvalues within each bin is computed. As such, the number of bins, $k$, is not dependent on the size of the network. Therefore, \newmethod can easily adapt to the evolving size of a dynamic graph. The main characteristics of \newmethod are:

\begin{itemize}[topsep=0pt,leftmargin=10pt]
    \item \textbf{Accurate:}  \newmethod achieves state-of-the-art performance in extensive synthetic experiments and can identify several major wars from the co-authorship network, MAG-History, of the History research community (while existing methods fail to adapt to the evolving size of this network).

    \item \textbf{Scalable:} \newmethod  has a linear time complexity with respect to the number of edges and is highly scalable. For example, on the MAG-History dataset with 2 million edges, \newmethod runs in 29 seconds on a stock laptop with CPU.
    
    \item \textbf{Attributed:} To the best of our knowledge, \newmethod is the first method to incorporate node attributes into change point detection for dynamic graphs. On our original COVID-flight dataset, \newmethod leverages the country code of airports~(nodes) to identify traffic disturbances due to flight restrictions specific to countries such as China and US.
\end{itemize}

\section{Related Work} \label{Sec:related}

In this section, we review methods for change point and event detection. We compare \newmethod to other approaches in Table~\ref{tab:salesman}. Note that current methods focus on graph structural anomalies, while \newmethod is the first method to incorporate node attributes and satisfies all the desired properties.

\begin{table}[t]  
\begin{center}
\caption{\newmethod is the only scalable method that detects both events and change points and also being the only method that accounts for attributes.\label{tab:salesman}} 
\begin{tabular}{ l|c|c|c|c|c|c }
       \diagbox{Method}{Property} &  \rotatebox{90}{Event} & \rotatebox{90}{Change Point} & \rotatebox{90}{Scalable} & \rotatebox{90}{Evolving Size} & \rotatebox{90}{Weights} & \rotatebox{90}{Attributes} \\ 
    \hline 
    Activity vector~\cite{ide2004eigenspace} & \CheckmarkBold & & \CheckmarkBold & \CheckmarkBold & \CheckmarkBold & \\ 
    TENSORSPLAT~\cite{koutra2012tensorsplat} & \CheckmarkBold & & &  & \CheckmarkBold & \\
    EdgeMonitoring~\cite{wang2017fast}       &      & \CheckmarkBold & \CheckmarkBold & &  &  \\
    SPOTLIGHT~\cite{eswaran2018spotlight} & \CheckmarkBold & &\CheckmarkBold & & \CheckmarkBold  & \\ 
    \method~\cite{huang2020laplacian}  & \CheckmarkBold & \CheckmarkBold &  & \CheckmarkBold & \CheckmarkBold & \\ 
    \hline
    \rowcolor{gray!20} \newmethod [this paper] & \CheckmarkBold  & \CheckmarkBold & \CheckmarkBold & \CheckmarkBold & \CheckmarkBold & \CheckmarkBold \\ 
    \hline
\end{tabular}
\end{center}
\vskip -0.2in
\end{table}

\subsection{Event Detection} 
Idé and Kashima~\cite{ide2004eigenspace} uses the principal eigenvector of the adjacency matrix to represent the graph at each snapshot~(called \emph{activity vector}). Koutra et al.~\cite{koutra2012tensorsplat} formulated dynamic graphs as high-order tensors and proposed to use the PARAFAC decomposition~\cite{bro1997parafac,harshman1970foundations} to obtain vector representations for anomaly scoring. SPOTLIGHT~\cite{eswaran2018spotlight} was proposed to spot anomalous graphs containing the sudden appearance or disappearance of large dense subgraphs. 

\subsection{Change Point Detection} 
Wang et al.~\cite{wang2017fast} modeled network evolution as a first-order Markov process and used MCMC sampling to design the EdgeMonitoring method. Recently, Huang et al.~\cite{huang2020laplacian} proposed Laplacian Anomaly Detection~(\method), which uses the exact singular values of the Laplacian matrix of each snapshot as the signature vector. \newmethod employs a similar anomaly detection pipeline to \method, while also utilizing spectral information from the Laplacian. Computing Singular Value Decomposition~(SVD) limits \method to small graphs while \newmethod can scale to millions of nodes and edges. 

\subsection{Network Density of States} 
Dong et al.~\cite{dong2019network} borrowed tools from condensed matter physics and added adaptations such as motif filtering to design an efficient approximation method for spectral density in large networks. Huang et al.~\cite{huang2021density} proposed a graph kernel which combines the local and global density of states of the normalized adjacency matrix for the graph classification task. ADOGE~\cite{sawlani2021fast} is an embedding method for exploratory graph analysis and graph classification on static graphs. To the best of our knowledge, our proposed \newmethod is the first method to model spectral density for dynamic graphs. 

\section{Problem Formulation and Notations}
\label{Sec:def}

We consider an undirected, weighted, dynamic graph $\mathbf{G}$ with node attributes (optional), as a sequence of graph snapshots, $\{\mathcal{G}_t \}_{t=1}^{T}$. Each $\mathcal{G}_t = (\mathcal{V}_t, \mathcal{E}_t, \mat{X}_t)$ represents the graph at time $t\in[1\dots T]$, where $\mathcal{V}_t$, $\mathcal{E}_t$ are the set of nodes and edges respectively, and  $\mat{X}_t \in \mathbb{R}^{|\mathcal{V}_t| \times N_a}$ is the attribute matrix,  where $N_a$ is the number of attributes. An edge $(i,j,w) \in \mathcal{E}_t$ connects node $i$ and node $j$ at time $t$ with weight $w$. We use  $\mat{A}_t \in \mathbb{R}^{|\mathcal{V}_t| \times |\mathcal{V}_t|}$ to denote the adjacency matrix of $\mathcal{G}_t$.

\subsection{Change Point Detection} 
The goal of change point detection is to identify anomalous time steps in a dynamic graph, i.e. snapshots with \emph{graph structures} that significantly deviate from the normal behavior. This requires an anomaly score function measuring the graph structural differences between the current snapshot and previously observed behaviors. In this work, we examine both \emph{events}, one time change to the graph structure and \emph{change points}, permanent alterations on the graph generative process. 

\subsection{Attribute Change Point Detection}
To the best of our knowledge, we are also the first work to incorporate node attributes in change point detection.
In addition to detecting change points in the graph structure, \emph{attribute} change point detection aims to identify time steps in which \emph{the alignment} between node attributes and graph structure deviates significantly from the norm. For example, in a network with communities, if the distribution of an attribute conditioned on the community drastically changes,  we say that an attribute change point has happened. 

\section{Scalable Change Point Detection}

\begin{algorithm}[t]
    \caption{\newmethod}
    \label{Algo:SCPD}
    \SetAlgoLined
    \KwIn{ \changed{$\mathbb{G} = \{ \mathcal{V}_t, \mathcal{E}_t, \mat{X}_t \}_{t=1}^{T}$ \& attribute $a$ (optional) }}
    \Parameter{$w_s,w_l$: short \& long sliding window sizes, 
                $N_z,N_m$: number of probe vectors \& Chebyshev moments}
    \KwOut{anomaly scores $Z^*$~(general) \& $Z_a^*$~(attribute)}
    \BlankLine
    {\footnotesize \color{gray}\tcc{{Compute Density of States}}}
    \ForEach{graph snapshot $\mathcal{G}_{t} = (\mathcal{V}_t, \mathcal{E}_t, \mat{X}_t) \in \mathbb{G}$}{
        Compute the symmetric normalized Laplacian $\mat{L}_{sym}$ \; 
        $\vec{\sigma}_t \leftarrow $ KPM\_DOS($\mat{L}_{sym}$, $N_z$,  $N_m$)\;
        \If{\changed{attribute $a$ is given} }{
        $\vec{\sigma}_{t,a} \leftarrow $ GQL\_LDOS($\mat{L}_{sym}$, $[\vec{X}_t]_{:,a}$, $N_z$,  $N_m$)\;}

        L2 normalize both $\vec{\sigma}_{t}$ and $\vec{\sigma}_{t,a}$\;
        
    }
    {\footnotesize \color{gray}\tcc{{Compute Anomaly Scores}}}
    \ForEach{time step $t$}{
    $\tilde{\sigma}_{t}^{w_s}$ and  $\tilde{\sigma}_{t}^{w_l}$ $\leftarrow$ left singular vector of context $\mat{C}_{t}^{w_s}$ and $\mat{C}_{t}^{w_l}$, where
    \begin{equation*} 
    \mathbf{C}^w_t = \left( \begin{array}{cccc}
                 |& | & &|  \\
                 \vec{\sigma}_{t-w-1} & \vec{\sigma}_{t-w-2} & \hdots & \vec{\sigma}_{t-1} \\ 
                 |& | & & |
            \end{array} \right) \in \mathbb{R}^{k \times w}
    \end{equation*}
    Obtain $Z^*$ score for $\vec{\sigma}_t$:  $Z =\max (1 - \vec{\sigma}_t{}^\top \tilde{\sigma}^{w_s}_t,1 - \vec{\sigma}_t{}^\top \tilde{\sigma}^{w_l}_t) $\;
     \If{d is given}{
     $\tilde{\sigma}_{t,a}^{w_s}$ and  $\tilde{\sigma}_{t,a}^{w_l}$ $\leftarrow$ left singular vector of context $\mat{C}_{t,a}^{w_s}$ and $\mat{C}_{t,a}^{w_l}$\;
    Obtain $Z_a^*$ score for $\vec{\sigma}_{t,a}$\;
    }
    }
    Return $Z^*, Z_a^*$ \;
\end{algorithm}

To detect anomalous snapshots, we embed each graph snapshot into a low-dimensional embedding called the signature vector based on the spectral density. Then, the normal behavior of the graph in the past is summarized into a vector. Lastly, we compare the signature from the current step with that of the past behavior and derive an anomaly score. For more details on \newmethod see Algorithm~\ref{Algo:SCPD}.

\subsection{Designing Signature Vector} 
Identifying change points require the comparison between multiple graph snapshots. In general, it is difficult to compare graphs directly, as shown in the graph isomorphism problem~\cite{weisfeiler1968reduction, xu2018powerful}. Therefore, we want to embed each graph snapshot into a low dimensional vector, called the \emph{signature vector}, and facilitate comparisons between vectors rather than graphs. In this work, we choose the (global) density of states~(DOS) of the Laplacian matrix as the signature vector, since it has the following desirable properties: 1.)\emph{scalable}, DOS can scale to graphs with millions of nodes and edges, 2.)\emph{independent of graph size}, DOS produces a fixed sized embedding independent of size of the graph, 3.) \emph{incorporates attributes}, the local DOS models the alignment between node attributes and eigenvectors thus can be used to model attribute change points. 

We use DOS to approximate the distribution of the Laplacian eigenvalues. The Laplacian eigenvalues capture many graph structures and properties~\cite{spielman2007spectral} and have shown strong empirical performance for anomaly detection~\cite{huang2020laplacian}. For example, the number of zero eigenvalues of the Laplacian matrix is equal to the number of connected components of the graph~\cite{von2007tutorial}, and the eigenvectors of the Laplacian matrix provide an effective way to represent a graph in a 2D plane~\cite{hall1970r}. In addition, the eigenvalues of the Laplacian and their multiplicity reflect the geometry of many fundamental graphs, such as  complete graphs, star graphs, and path graphs. However, computing all Laplacian eigenvalues of a graph requires $O(|\mathcal{V}| \cdot |\mathcal{E}|)$, which is only practical for small graphs, while DOS can be computed for graphs with millions of nodes. Later in this section, we show how to compute DOS efficiently, and in Section~\ref{Sec:synthetic} and~\ref{Sec:real}, we demonstrate that DOS has state-of-the-art performance in change point detection.

\subsection{Computing Anomaly Score} 

After computing the signature vectors for each timestamp, now we explain how to detect anomalous snapshots. We assume that when an anomaly arrives, it will be significantly different from recent snapshots. Therefore, we extract the ``expected'' or ``normal'' behavior of the dynamic graph from a context window of size $w$ from the past $w$ signature vectors. To obtain unit vectors, $L2$ normalization is performed on the set of the signature vectors $\vec{\sigma}_{t-w-1}, \hdots, \vec{\sigma}_{t-1}$. Then, we stack the normalized vectors to form the context matrix $\mathbf{C}^w_t \in \mathbb{R}^{k \times w}$ of time t, where $k$ is the length of the signature vector. We compute the left singular vector of $\mathbf{C}^w_t$ to be the summarized normal behavior vector $\tilde{\sigma}_t^w$ (which can be seen as a weighted average over the context window). A smaller context window can detect more sudden or abrupt changes, while a longer window can model gradual and continuous changes. Therefore, we use a short window with size $w_s$ and a long window with size $w_l$ to detect both events and change points.

Now we can compute the anomaly score  at time $t$ as 
\begin{equation*}
Z_t = 1 - \frac{\vec{\sigma}_t^\top \tilde{\sigma}_t^w }{ \|\vec{\sigma}_t\|_2 \|\tilde{\sigma}_t^w\|_2} = 1 - \vec{\sigma}_t^\top \tilde{\sigma}_t^w = 1 - \cos{\theta}
\end{equation*}
where $\cos{\theta}$ is the cosine similarity between the current signature vector $\vec{\sigma}_t$ and the normal behavior vector $\tilde{\sigma}_t^w$. In this way, $Z \in [0,1]$ and when $Z$ is closer to 1, the current snapshot is significantly different from the normal behavior, thus more likely to be an anomaly. The $Z$ scores from windows of size $w_s$ and $w_l$ are then aggregated by the max operation. To emphasize the increase in anomaly score, we compute the difference in anomaly score with the previous step using $Z_t^* = \min(Z_t - Z_{t-1},0)$. Finally, the points with the largest $Z^*$ are selected as anomalies. We show the $Z^*$ score in all figures in this work. 


\subsection{Approximating Spectral Density} For clarity, we drop the $t$ subscript in this section. The Laplacian matrix $\mat{L} \in \mathbb{R}^{|\mathcal{V}| \times| \mathcal{V}|}$ is defined as $\mat{L} = \mat{D} - \mat{A}$ where $\mat{D} \in \mathbb{R}^{|\mathcal{V}| \times |\mathcal{V}|}$, $\mat{A} \in \mathbb{R}^{|\mathcal{V}| \times| \mathcal{V}|}$ are the diagonal degree matrix and the adjacency matrix. In this work, we use the symmetric normalized Laplacian $\mat{L}_{sym}$, defined as,
\begin{equation*}
\mat{L}_{sym} = \mat{D}^{-\frac{1}{2}} \mat{L} \mat{D}^{-\frac{1}{2}} = \mat{I} - \mat{D}^{-\frac{1}{2}} \mat{A} \mat{D}^{-\frac{1}{2}}
\end{equation*}
to present the graph at each snapshot. Consider the eigendecomposition of $\mat{L}_{sym} = \mat{Q} \mat{\Lambda} \mat{Q}^T$ where $\mat{\Lambda} =  diag(\lambda_1, \hdots, \lambda_{| \mathcal{V}|})$ and $\mat{Q} = [\vec{q}_1, \dots, \vec{q}_{| \mathcal{V}|}]$ is an orthogonal matrix. We can now define the Density of States or the spectral density as,

\begin{definition}[Density of States~(DOS)]
The global density of states or spectral density induced by $\mat{L}_{sym}$ is: 
\begin{equation}\label{eq:def.dos}
    \mu(\lambda) = \frac{1}{| \mathcal{V}|} \sum_{i=1}^{| \mathcal{V}|}\delta(\lambda-\lambda_i)
\end{equation} where $\delta$ is the Dirac delta function and $\lambda_i$ is the $i$-th eigenvalue.
\end{definition}%
Intuitively, $\mu(\lambda)$ measures the portion of eigenvalues that are equal to $\lambda$. In practice, we discretize the range of $\lambda$ into equal sized intervals and approximate how many $\lambda$s falls within each interval. Therefore, across all intervals, the shape of the distribution of eigenvalues are approximated. We use the Kernel Polynomial Method~(KPM)~\cite{dong2019network} to approximate the density function through a finite number polynomial expansion in the dual basis of the Chebyshev basis, and the spectrum is adjusted to be in $[-1,1]$ for numerical stability.  

To incorporate attributes, we also consider the Local Density of States:
\begin{definition}[Local Density of States~(LDOS)]
For any given input vector $\vec{v} \in \mathbb{R}^{N}$, the local density of states is: 
\begin{equation}\label{eq:def.ldos}
    \mu(\lambda ; \vec{v}) = \sum_{i=1}^{{| \mathcal{V}|}} |\vec{v}^T \vec{q}_i|^2 \delta(\lambda-\lambda_i)
\end{equation} where $\vec{v}$ is an input vector, and $\lambda_i$ and $\vec{q}_i$ are the $i$-th eigenvalue and eigenvector, respectively. 
\end{definition}
The term $\vec{v}^T \vec{q}_i$ acts as a weight on the $i$th bin of the spectral histogram. To incorporate node attributes, we set $\vec{v} = \vec{x}$ where $\vec{x} \in \mathbb{R}^{| \mathcal{V}|}$ is an attribute vector. This can be interpreted as the \emph{alignment} between the node attribute vector and the graph structure of the group of nodes with such attribute. As the alignment is measured in each eigenvalue interval~(similarly to DOS), we obtain a LDOS embedding of size $k$ for each attribute and each possible category. By tracking this embedding over time, the anomalous evolution specific to the given attribute is captured. Here, categorical attributes are one-hot encoded, and numerical attributes are normalized by the sum. The Gauss Quadrature and Lanczos~(GQL)~\cite{dong2019network} method to approximate the LDOS with attribute vectors.

\subsection{Computational Complexity} For unattributed dynamic graphs, \newmethod has the complexity of $O(N_z \cdot N_m \cdot |\mathcal{E}|)$ for a given snapshot with $|\mathcal{E}|$ edges. $N_z$ and $N_m$ are hyperparameters in the KPM computation representing the number of probe vectors and Chebychev moments, respectively. For all experiments, we set $N_z = 100$, $N_m = 20$. We also use $k=50$ equal sized bins to divide up the range of eigenvalues. We show that \newmethod is robust to hyperparameter choices in Appendix~\ref{app:sensitivity}.

For attributed dynamic graphs, we use the GQL method to compute LDOS for attribute change point detection. GQL method performs the eigendecomposition of a tridiagonal matrix with $O(|\mathcal{V}|^2)$ worst case complexity. Note that in practice, such computation is very fast~\cite{sawlani2021fast}. Therefore, for a given attribute on a dynamic graph, \newmethod's time complexity is $O(\eta \cdot |\mathcal{E}| + |\mathcal{V}|^2)$ for a given snapshot. In practice, \newmethod is very fast, only costing $5$ seconds to run on the COVID flight network with close to 1 million edges and $5$ node attributes with an AMD Ryzen 5 1600 Processor and 16GB memory.

\section{Synthetic Experiments} \label{Sec:synthetic}

In this section, we conduct experiments with the Stochastic Block Model~(SBM)~\cite{holland1983stochastic} and the Barabási–Albert~(BA) model~\cite{barabasi2013network} as synthetic graph generators and plant 7 ground truth anomalies for all experiments. We report the Hits$@n$ metric same as in~\cite{huang2020laplacian} and the execution time over 5 trials. We discuss our considered baselines and contenders in Section~\ref{sub:contend}, evaluation setting in Section~\ref{sub:eval}, and planted anomalies in Section~\ref{sub:planted}. 

\subsection{Contenders and Baselines} \label{sub:contend}
We compare \newmethod with the current state-of-the-art baselines~(see Section~\ref{Sec:related} for more details). The same short-term and long-term window sizes~($s=5$ and $l=10$) are used if applicable. For methods with stochasticity, we report the average result over 5 runs. 
\begin{itemize}[align=parleft,left=0pt..1em]
    \item \textbf{SPOTLIGHT~\cite{eswaran2018spotlight} with \href{https://github.com/kLabUM/rrcf}{RRCF}~\cite{guha2016robust}}: SPOTLIGHT is a randomized sketching-based approach to detect anomalous graphs containing sudden appearance or disappearance of large dense subgraphs. We implemented SPOTLIGHT from scratch in Python and then used the recommended anomaly detection method Robust Random Cut Forest~(RRCF)~\cite{guha2016robust} for the SPOTLIGHT embeddings. Following~\cite{eswaran2018spotlight}, we set $p=q=0.2$ and $k=50$ sketch dimensions. 
    
    \item \textbf{SPOTLIGHT~\cite{eswaran2018spotlight} with a sum, or SPOTLIGHTs}: to achieve a more competitive performance with SPOTLIGHT for change point detection, we introduce a simple anomaly detection pipeline to use with the computed SPOTLIGHT embeddings. We find that simply summing over all SPOTLIGHT dimensions in each snapshot is a good proxy for the anomaly score, and we use the difference between the sum score of $t$ and $(t-1)$ as the final anomaly score.  
    
    \item \textbf{LAD~\cite{huang2020laplacian}}: Laplacian Anomaly Detection~(\method) utilizes all the eigenvalues of the Laplacian matrix as the embedding vector for each snapshot. For \method, we use the symmetric normalized Laplacian $\mat{L}_{sym}$ instead of unnormalized Laplacian $\mat{L}_t$ because $\mat{L}_{sym}$ improved \method performance across both BA and SBM datasets. Using the recommendation from~\cite{huang2020laplacian}, we compute all the eigenvalues in the Laplacian spectrum to achieve the best performance for \method. Due to high cost of SVD, we limit the computational time to less than 5 hrs, and if the computation is not finished within such time limit, we report N/A as it is considered not scalable.
    
    \item \textbf{EdgeMonitoring~\cite{wang2017fast}}:  EdgeMonitoring uses joint edge probabilities as the feature vector and models network evolution as a first-order Markov process. We use the official MATLAB implementation kindly provided by the authors of EdgeMonitoring and sample 250 edges in 25 equal-sized groups as in~\cite{wang2017fast}. The detected anomalies are based on the Euclidean Distance. 
    
\end{itemize}

\subsection{Performance Evaluation} \label{sub:eval}
We conduct experiments with the Stochastic Block Model~(SBM) and the Barabási–Albert~(BA) model as synthetic graph generators and plant ground truth anomalies for all experiments. Similar to~\cite{huang2020laplacian}, we use Hits$@n$ metric, which reports the number of correctly detected anomalies out of the top $n$ steps with the highest anomaly scores. For synthetic experiments, for uniformity, we plant 7 anomalies and report Hits$@7$~(which penalizes both false positive and false negative anomaly alarms). We also report the execution time in seconds on a desktop with AMD Ryzen 5 1600 CPU and 16 GB memory. We use the ground truth labels from the generation process for evaluation.

\subsection{Planted Anomaly Details} \label{sub:planted}
In this section, we describe the details of the planted anomalies and the random graph generators used in Section~\ref{Sec:synthetic}. Table~\ref{tab:SBM} a). and b). describes the anomalies in the SBM hybrid experiment and SBM attribute experiment, respectively. Table~\ref{tab:evolveBA} provides details for the anomalies in the SBM evolving size experiment and the BA experiment. 

SBM~\cite{holland1983stochastic} is a widely used graph generation model for community structures. The key parameters of the SBM model are: 1). the partitioning of communities, 2). the intra-community connectivity $p_{in}$, and 3). the cross-community connectivity $p_{out}$. For simplicity, we assume equal-sized communities and instead focus on changing the number of communities $N_c$. $p_{in}$ and $p_{out}$ determine the probability of an edge existing between nodes of the same community and different communities, respectively. They also control the sparsity of the dynamic graph.

\definecolor{orange}{RGB}{222,119,174}
\definecolor{purple}{RGB}{53,151,143}

\begin{table*}[t!]
\caption{Experiment Setting: the changes in the generative model in \\
(a) the setting for the SBM hybrid experiment where both events and change points occur in a SBM model, $N_c$ is the number of equal-sized communities, $p_{in}$ is the intra-community connectivity and $p_{out}$ is the cross-community connectivity.\\
(b) the setting for the SBM attribute experiment where change points in the node attribute and graph structure of a SBM model occurs}\label{tab:SBM}
\begin{subtable}[t]{0.5\textwidth}
\begin{tabular}{c | c | l | l | l}
\toprule
\multicolumn{5}{c}{SBM Change Points Details}\\
\midrule
Time & Type & $N_c$ & $p_{in}$ & $p_{out}$\\
\midrule
0 & start point & 4 &  0.030 &  0.005 \\\rowcolor{orange!50}
16 & event & 4 &  0.030 & \textbf{0.015} \\\rowcolor{purple!50}
31 & change point & \textbf{10} &  0.030 &  0.005 \\\rowcolor{orange!50}
61 & event & 10 &  0.030 & \textbf{0.015} \\\rowcolor{purple!50}
76 & change point & \textbf{2} &  0.030 &  0.005 \\\rowcolor{orange!50}
91 & event & 2 &  0.030 & \textbf{0.015} \\\rowcolor{purple!50}
106 & change point & \textbf{4} &  0.030 &  0.005 \\\rowcolor{orange!50}
136 & event & 4 &  0.030 & \textbf{0.015} \\
\bottomrule
\end{tabular}
\caption{anomalies in the SBM hybrid experiment.}
\end{subtable}%
\begin{subtable}[t]{0.5\textwidth}
\begin{tabular}{c | c | l | l}
\toprule
\multicolumn{4}{c}{SBM Node Attribute and Structural Changes}\\
\midrule
Time & Type & $N_c$ & Node Attributes\\
\midrule

0 & start point & 4 & homogeneous \\\rowcolor{red!50}
16 & change point & 4 & \textbf{heterogeneous} \\\rowcolor{purple!50}
31 & change point & \textbf{10} & heterogeneous \\\rowcolor{red!50}
61 & change point & 10 & \textbf{homogeneous} \\\rowcolor{purple!50}
76 & change point &\textbf{2} & homogeneous \\\rowcolor{red!50}
91 & change point & 2 & \textbf{heterogeneous} \\\rowcolor{purple!50}
106 & change point & \textbf{4} & heterogeneous \\\rowcolor{red!50}
136 & change point & 4 & \textbf{homogeneous} \\
\bottomrule
\end{tabular}
\caption{anomalies in the SBM attribute experiment}
\end{subtable}%

\end{table*}

\begin{table*}[t!]
\caption{Experiment Setting: the changes in the generative model in \\
(a) setting for the SBM hybrid experiment where both events and change points occur in a SBM model, $N_c$ is the number of equal-sized communities, $p_{in}$ is the intra-community connectivity and $p_{out}$ is the cross-community connectivity.\\
(b) setting for the BA experiment where the parameter $m$ is the number of edges to attach from a new node to existing nodes. Higher $m$ value = higher color intensity.}\label{tab:evolveBA}

\begin{subtable}[t]{0.6\textwidth}
\begin{tabular}{c | l | c | c}
\toprule
\multicolumn{4}{c}{SBM Anomaly Details}\\
\midrule
Time & Community Sizes & $p_{in}$ & $p_{out}$ \\
\midrule
0 & 300,300 & 0.030 & 0.005 \\ 
16 & 300,300,\textbf{300} & 0.030 & 0.005 \\
31 & 300,300,\textbf{300,300} & 0.030 & 0.005 \\
61 & 300,300,\textbf{150,150,150,150} & 0.030 & 0.005 \\
76 & 300,300,\textbf{300,300} & 0.030 & 0.005 \\
91 & \textbf{150,150,150,150},300,300 & 0.030 & 0.005 \\
106 & \textbf{300,300},300,300 & 0.030 & 0.005 \\
136 & 300,300,300,300 & 0.030 & \textbf{0.015}\\
\bottomrule
\end{tabular}
\caption{anomalies in SBM Evolving Size Experiment}
\end{subtable}%
\begin{subtable}[t]{0.4\textwidth}
\begin{tabular}{c | l | c}
    \toprule
    \multicolumn{3}{c}{BA Anomaly Details}\\
    \midrule
    Time & Type & $m$\\
    \midrule
    0 & start point &  1\\\rowcolor{green!2}
    16 & change point & 2\\\rowcolor{green!4}
    31 & change point & 3\\\rowcolor{green!7}
    61 & change point & 4\\\rowcolor{green!10}
    76 & change point & 5\\\rowcolor{green!13}
    91 & change point & 6\\\rowcolor{green!17}
    106 & change point & 7\\\rowcolor{green!20}
    136 & change point & 8\\
    \bottomrule
\end{tabular}
\caption{anomalies in the BA experiment.}
\end{subtable}%
\end{table*}


\subsection{SBM Hybrid Experiment} We follow the Hybrid setting in~\cite{huang2020laplacian}. SBM~\cite{holland1983stochastic} is used to generate equal-sized communities with $p_{in}$ being the intra-community connectivity and $p_{out}$ being the cross-community connectivity. Change points are the merging or splitting of communities in the dynamic graph, and events are one-time boosts in cross-community connectivity $p_{out}$. Figure~\ref{Fig:DOS.approx} shows that \newmethod perfectly identifies all the events and change points on a dynamic SBM graph. We also visualize the signature vectors~(the computed DOS or distribution of eigenvalues) as a heatmap. The events~(time point 16,61,91,136) correspond to an energetic burst in the signature vector. And the change points correspond to the shifts in the distribution of Laplacian eigenvalues. Interestingly, the width of the distribution seems to correlate with the number of communities $N_c$.

\subsection{SBM Attribute Experiment} We want to demonstrate \newmethod's ability to detect anomalous evolution of the node attributes in a dynamic graph. A SBM model is used to construct communities for nodes while each node has a binary attribute. The attributes within a community can be either \emph{homogeneous} or \emph{heterogeneous}. In a homogeneous community, all nodes have the same attribute, while half of all communities have label one and the other half have label two. In a heterogeneous community, each node has $0.5$ probability of being either one or two, and the node attribute is no longer dependent on community structure. The change points are time points where the node attributes change to \emph{homogeneous} or \emph{heterogeneous}. \newmethod is able to recover all change points~(16,61,91, and 136) related to node attributes and detect both the change from homogeneous communities to heterogeneous ones as well as the reverse. 

Interestingly, by just using LDOS embedding, \newmethod also captures two out of three structural change points when the number of communities changes. This is because, in Equation~\ref{eq:def.ldos}, the $\vec{v}^T \vec{q}_i$ term measures the alignment between the input attribute vector and the eigenvector corresponding to $i$th eigenvalue of the Laplacian. This term would change either due to a change in $\vec{v}$ or $\vec{q}_i$ or both. In this case, the change in community structure also caused a change in the eigenvectors of the Laplacian. Note that as none of the alternative methods take into account the node attributes, \newmethod is the only method which can detect the anomalous evolution of node attributes.

\subsection{SBM Evolving Size Experiment}

\begin{figure}[t]
\centering
\begin{subfigure}{.45\textwidth}
  \centering
  \includegraphics[width=\linewidth]{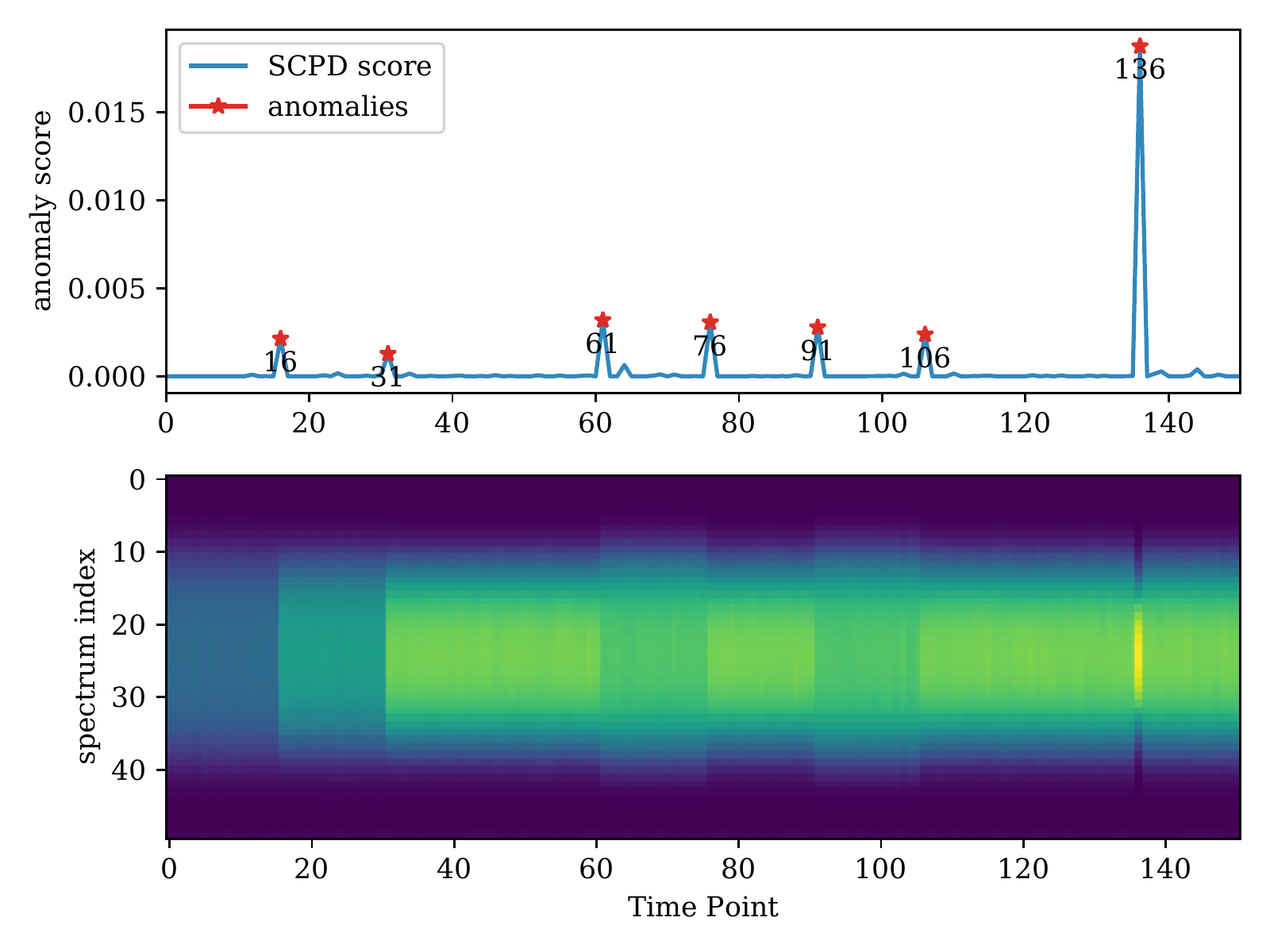}
  \caption{SBM evolving size experiment}
  \label{Fig:SBM_evolve}
\end{subfigure}%
\begin{subfigure}{.5\textwidth}
  \centering
  \includegraphics[width=\linewidth]{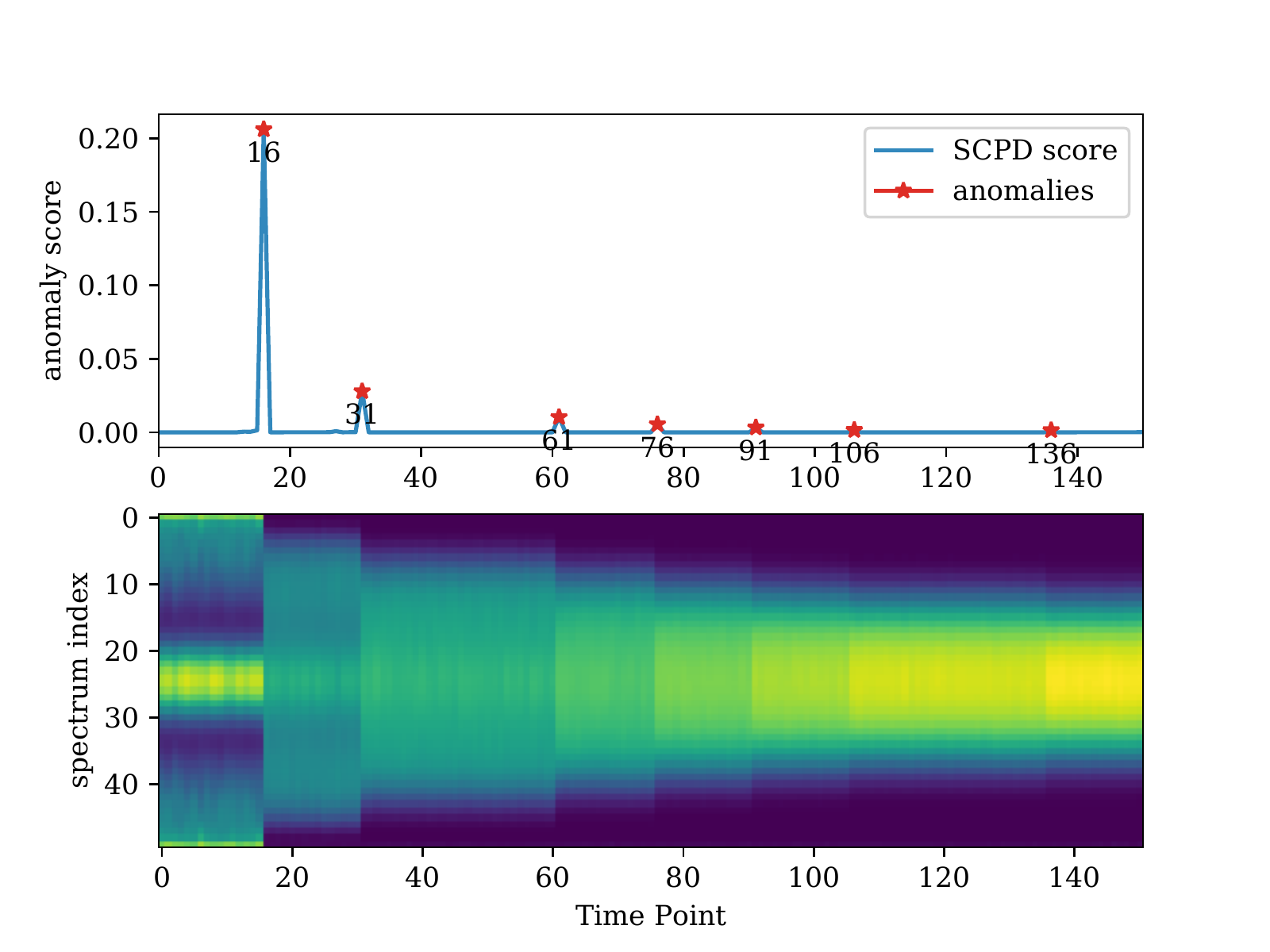}
  \caption{The BA experiment} 
  \label{Fig:dos.AB}
\end{subfigure}
\caption{
a). \newmethod perfectly captures anomalies in the SBM evolving size experiment \\ 
b). \newmethod perfectly recovers all  \underline{change points} for the BA experiment.}
\label{Fig:syn_exp}
\vskip -0.2in
\end{figure}

We examine \newmethod's ability to adapt to the evolving size of a dynamic graph~(with a SBM as the graph generator).  Initially, there are two communities with 300 nodes each. Later on, additional nodes are added and forming a total of 4 communities. Some change points involve only nodes from the initial step, while some involve only newly added nodes. Figure~\ref{Fig:SBM_evolve} shows that \newmethod is able to adapt to dynamic graphs with evolving sizes and correctly predicts anomalies in the global graph structure. 
Only \newmethod and \method are able to correctly detect all anomalies, while SPOTLIGHT and EdgeMonitoring can only detect changes local to the initial set of nodes. This shows that \newmethod can effectively adapt to the evolving size of dynamic graphs.


\subsection{BA Experiment} 

We evaluate \newmethod performance in a different graph distribution, the BA model. In this experiment, the change points correspond to the densification of the network~(parameter $m$, increased number of edges attached from a new node to an existing node). From Figure~\ref{Fig:dos.AB}, we see that \newmethod is able to detect all change points in the BA model and the most drastic change in DOS happens when $m$ changes from one to two and the graph becomes connected. This is because the number of zero eigenvalues in the Laplacian matrix corresponds to the number of connected components in the graph; thus, when the graph is connected, the smallest eigenvalue intervals become less energetic. 

\begin{table*}[t!]
    \centering
    \caption{\newmethod can efficiently operate on large graphs while achieving the state-of-the-art performance. Each dynamic graph has 151 time steps. The results are Hits@$7$ averaged over 5 trials and the mean and standard deviations are reported. We consider a method not applicable (N/A) if the computation takes longer than 5 days.}
    \resizebox{\textwidth}{!}{
    \begin{tabular}{l | c | c | c | c | c}
    \toprule
    Generator & \multicolumn{3}{c|}{SBM} & \multicolumn{2}{c}{BA} \\
    \midrule
    Experiment & \multicolumn{2}{c|}{Hybrid} & Evolving Size & \multicolumn{2}{c}{Change Point} \\
    \midrule
    Total Edges~(millions) & 0.8 m & 56.9 m & 1.0 m & 0.6 m & 5.5 m \\
    \midrule
    \rowcolor{gray!20}\newmethod~(ours) & \textbf{1.00 $\pm$ 0.00} & \textbf{1.00 $\pm$ 0.00} & \textbf{1.00 $\pm$ 0.00} & \textbf{1.00 $\pm$ 0.00} &  \textbf{1.00 $\pm$ 0.00}\\
    
    \method~\cite{huang2020laplacian} & \textbf{1.00 $\pm$ 0.00} & N/A & \textbf{1.00 $\pm$ 0.00} & \textbf{1.00 $\pm$ 0.00} &  N/A\\
    
    SPOTLIGHT~\cite{eswaran2018spotlight} & $0.31 \pm 0.06$ & $0.57 \pm 0.00$ & $0.20 \pm 0.07$ & $0.06 \pm 0.07$ &  $0.11 \pm 0.11$ \\
    
    SPOTLIGHTs & $0.71 \pm 0.00$ &  $0.71 \pm 0.00$ & $0.31 \pm 0.06$ & \textbf{1.00 $\pm$ 0.00} & \textbf{1.00 $\pm$ 0.00}\\ 
    
    EdgeMonitoring~\cite{wang2017fast} & 0.06 $\pm$ 0.11 & 0.00 $\pm$ 0.00 &  0.14 $\pm$ 0.00 & 0.06 $\pm$ 0.07 & 0.17 $\pm$ 0.11\\
    \bottomrule
    \end{tabular}}
    \label{tab:results}
    \vskip -0.1in
\end{table*}

\begin{figure}[t]
    \begin{center}
        \centerline{\includegraphics[width=0.6\textwidth]{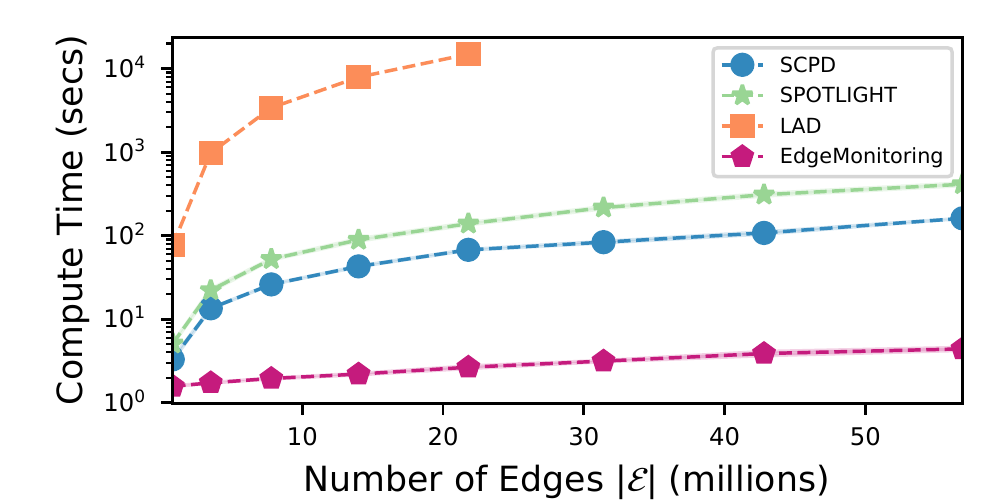}}
        \caption{Compute time comparison between different methods on the SBM hybrid experiment with varying number of edges.}
        \label{Fig:time}
    \end{center}
    \vskip -0.4in
\end{figure}

\subsection{Summary of Results} 

Table~\ref{tab:results} compares the performance of \newmethod with state-of-the-art methods on synthetic experiments. The SBM attribute experiment is not included, as only \newmethod can incorporate node attributes. The considered baselines include LAD~\cite{huang2020laplacian}, SPOTLIGHT~\cite{eswaran2018spotlight} and EdgeMonitoring~\cite{wang2017fast}. The original SPOTLIGHT (with RRCF~\cite{guha2016robust} detector) and our own variant, SPOTLIGHT with sum predictor, called SPOTLIGHTs are both included. Across all experiments, \newmethod has the best overall performance. With the default RRCF anomaly detection pipeline, SPOTLIGHT~\cite{eswaran2018spotlight} performs poorly on the BA model and middling performance on the SBM hybrid experiment. With the simple sum predictor introduced by us, SPOTLIGHTs is a much closer competitor with a strong performance on the BA model and improved performance on the SBM hybrid experiment. However, SPOTLIGHTs is still not able to detect changes in the evolving size experiment and overall outperformed by \newmethod. EdgeMonitoring~\cite{wang2017fast} has low performance in the synthetic experiments due to its dependency on node ordering as well as the assumption that only a small percentage of edges would be resampled in a dynamic graph. The closest competitor to \newmethod is \method~\cite{huang2020laplacian}. However, computing all the eigenvalues in \method is prohibitively expensive on large graphs and thus reported as not applicable.


In Figure~\ref{Fig:time}, we compare the computational time across different methods in the SBM hybrid experiment. The most expensive is \method as it has worst-case complexity cubic to the number of nodes, thus having a poor trade-off between performance and efficiency. In contrast, both \newmethod and SPOTLIGHT have complexity linear to the number of edges. However, \newmethod outperforms SPOTLIGHT across all experiments shown in Table~\ref{tab:results}. Therefore, \newmethod has the best trade-off between compute time and performance. Lastly, EdgeMonitoring has sublinear complexity to the number of edges; however, its performance is not ideal. 

In Appendix~\ref{app:sensitivity}, we show that \newmethod is robust to the choice of hyperparameters, including the number of probing vectors $N_z$, the number of Chebychev moments $N_m$, and the number of equal-sized bins $k$ in the range of Laplacian eigenvalues.

\section{Real World Experiments} \label{Sec:real}

We empirically evaluate \newmethod on two real-world dynamic networks and cross-reference anomalies detected by \newmethod with significant events. We provide more dataset details in Appendix~\ref{app:real}. We also report the results for SPOTLIGHT and EdgeMonitoring on these datasets in Appendix~\ref{app:baseline_real}. 


\subsection{MAG History Co-authorship Network}
MAG-History is a co-authorship dynamic network extracted from the Microsoft Academic Graph~(MAG)~\cite{Sinha-2015-MAG,Benson-2018-simplicial} by identifying publications which are marked with the "History" tag. The processed dataset is an undirected dynamic graph from 1837 to 2018. There are 2.8 million projected edges across all time steps and 0.7 million nodes in total. To compute the DOS embedding for this dataset, \newmethod only takes 30 seconds.

\begin{figure*}[t]
    \begin{center}
        \centerline{\includegraphics[width=0.9\textwidth]{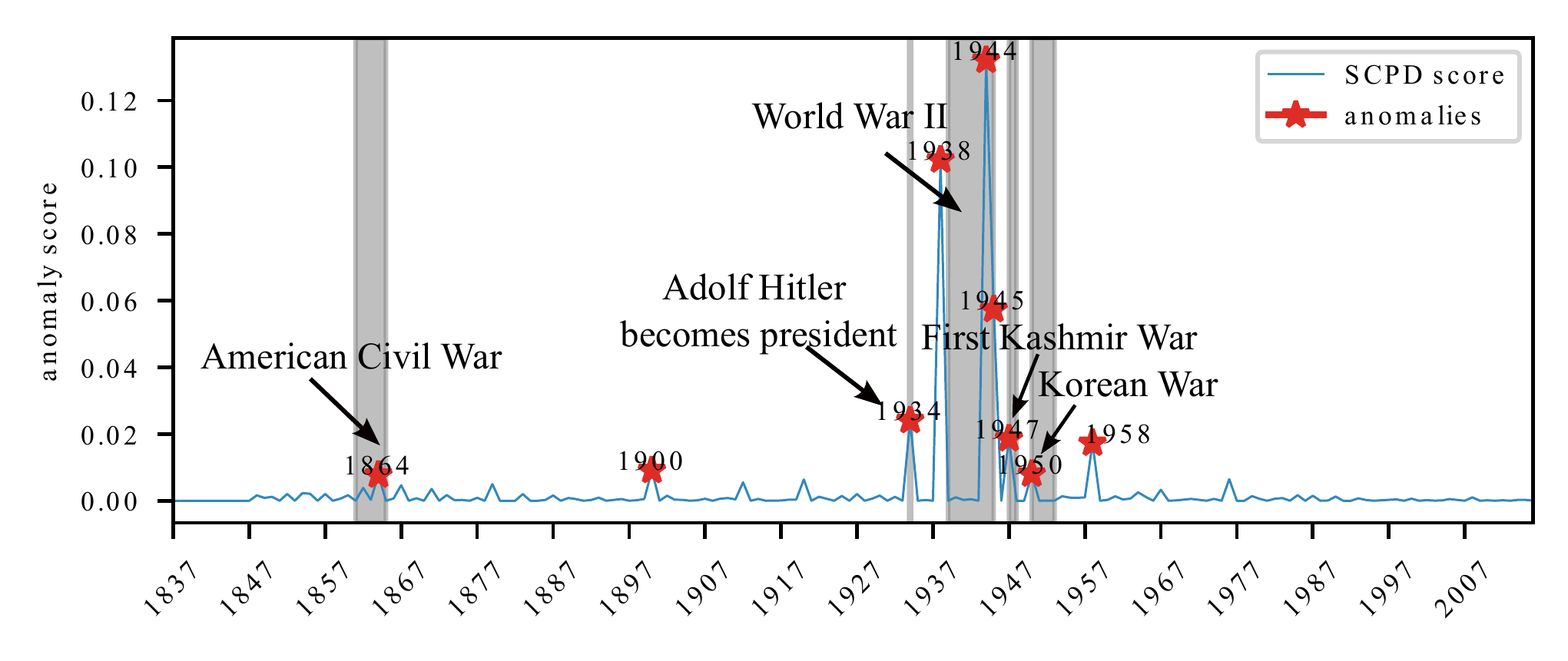}} 
        \caption{\newmethod detects significant historical events from the MAG-History dataset.}
        \label{Fig:history.gdos}
    \end{center}
    \vskip -0.3in
\end{figure*}


Figure~\ref{Fig:history.gdos} shows the anomalies detected by \newmethod. Interestingly, many of the anomalies correspond to important historical events such as the American Civil War~(1861-1865), Adolf Hitler's rise to power~(1934), Second World War~(1939-1945), First Kashmir War~(1947-1948) and Korean War~(1950-1953). The relation between the change in co-authorship graph structure and these historical events can be an interesting direction for future work. In comparison, both variants of SPOTLIGHT miss the second world war as a top anomaly, while EdgeMonitoring's output is noisy, and many data points share high anomaly scores.


\begin{figure}[t]
\centering
\begin{subfigure}{.5\textwidth}
  \centering
  \includegraphics[width=\linewidth]{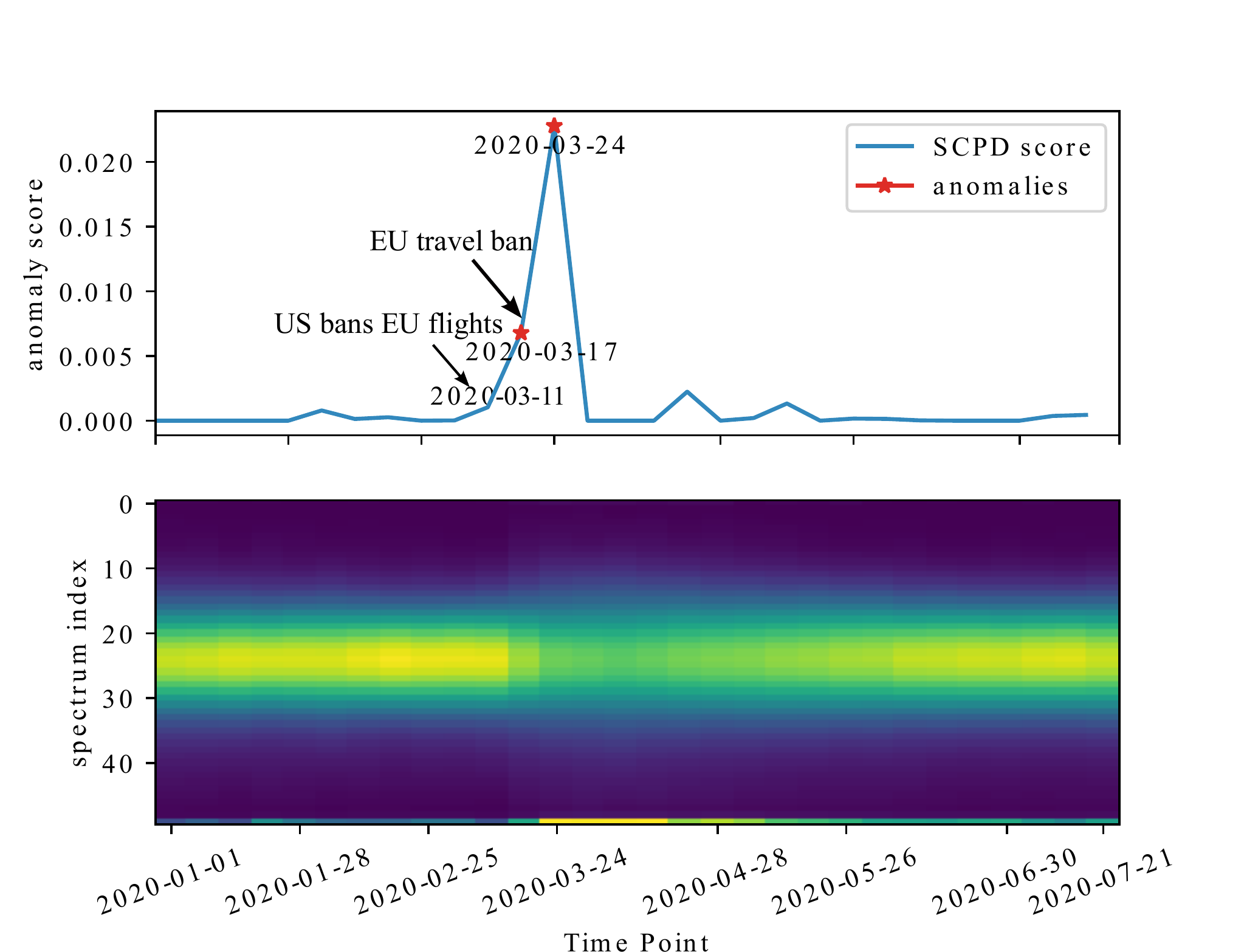}
  \caption{Graph structural anomalies}
  \label{Fig:covid.gdos}
\end{subfigure}%
\begin{subfigure}{.5\textwidth}
  \centering
  \includegraphics[width=\linewidth]{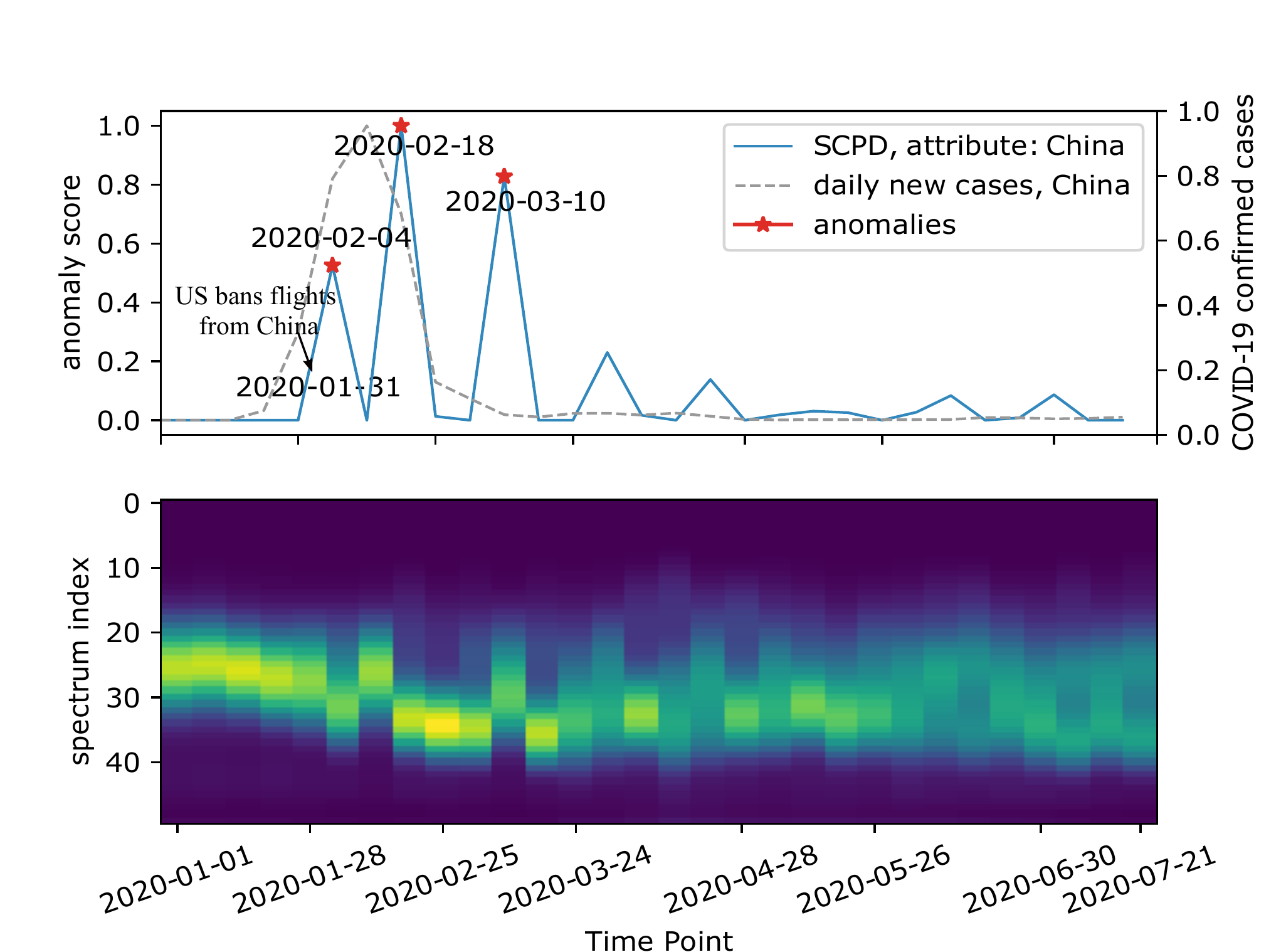}
  \caption{Node attribute anomalies} 
  \label{Fig:covid.china}
\end{subfigure}
\caption{
a). \newmethod detects the week of 03.17 and 03.24 as structural anomalies in the global flight network in 2020. On 03.17, the European Union closed its borders to travellers, thus causing wide spread disruption in international flights. \\ 
b). \newmethod detects the closure of flight routes to China due to COVID interventions at the beginning of Feb 2020. The anomaly score and case numbers are normalized. 
}
\label{Fig:flight}
\vskip -0.2in
\end{figure}

\subsection{COVID Flight Network}
the COVID flight network~\footnote{\url{https://zenodo.org/record/3974209/\#.Yf62HepKguU}}~\cite{schafer2014bringing, olive2019traffic} is a dynamic air traffic network during the COVID-19 pandemic. The nodes are airports, and each edge is an undirected timestamped tracked flight with the frequency as edge weight. We examine the period from 01-01-2020 to 07-27-2020. We use a full week as the duration of each snapshot to reduce the noise and variability from daily flights. 
Figure~\ref{Fig:covid.gdos} shows the graph structural anomalies detected by \newmethod using the DOS embeddings as signature vectors. The two weeks with the highest anomaly scores are 03-17-2020 and 03-24-2020. On 03-17, the European Union adopted a 30-day ban on non-essential travel to at least 26 European countries from the rest of the world~(see \href{https://www.nytimes.com/article/coronavirus-timeline.html}{here}). On 03-11, the US President banned travel from 26 European countries. These events are detected by \newmethod as travel bans severely disrupt the international flight network. In comparison, SPOTLIGHT detects the week of 02-11 corresponding to flight restrictions on China, while EdgeMonitoring also detects mid March as anomalies. 

Figure~\ref{Fig:covid.china} shows \newmethod's detected anomalies when the node attribute is set to be an indicator vector for which nodes are Chinese airports. The detected anomalies lie mainly in February and early March because the COVID outbreak was first detected in China in January 2020. On 01-31-2020, the Trump administration suspended entry into the United States by any foreign nationals who had traveled to China in the past 14 days~(see \href{https://www.nytimes.com/article/coronavirus-timeline.html}{here}). Therefore, the anomaly observed by \newmethod on the week of 02-04-2020 is likely the directed result of the imposed travel restriction. Note that Figure~\ref{Fig:covid.china} shows that the peak of new daily cases in China~\footnote{\url{https://www.worldometers.info/coronavirus/country/china/}} corresponds to a peak in anomaly score, likely because of reduced domestic and international flights at that time. For Canada, \newmethod also detects the implementation of travel restrictions~(see Appendix~\ref{app:covid}). Therefore, \newmethod captures both the structural anomalies in the flight network as well as anomalies specific to the set of nodes with the same attribute. 


\subsection{Stablecoin Transaction Network}

\begin{figure*}[t]
    \begin{center}
        \centerline{\includegraphics[width=\textwidth]{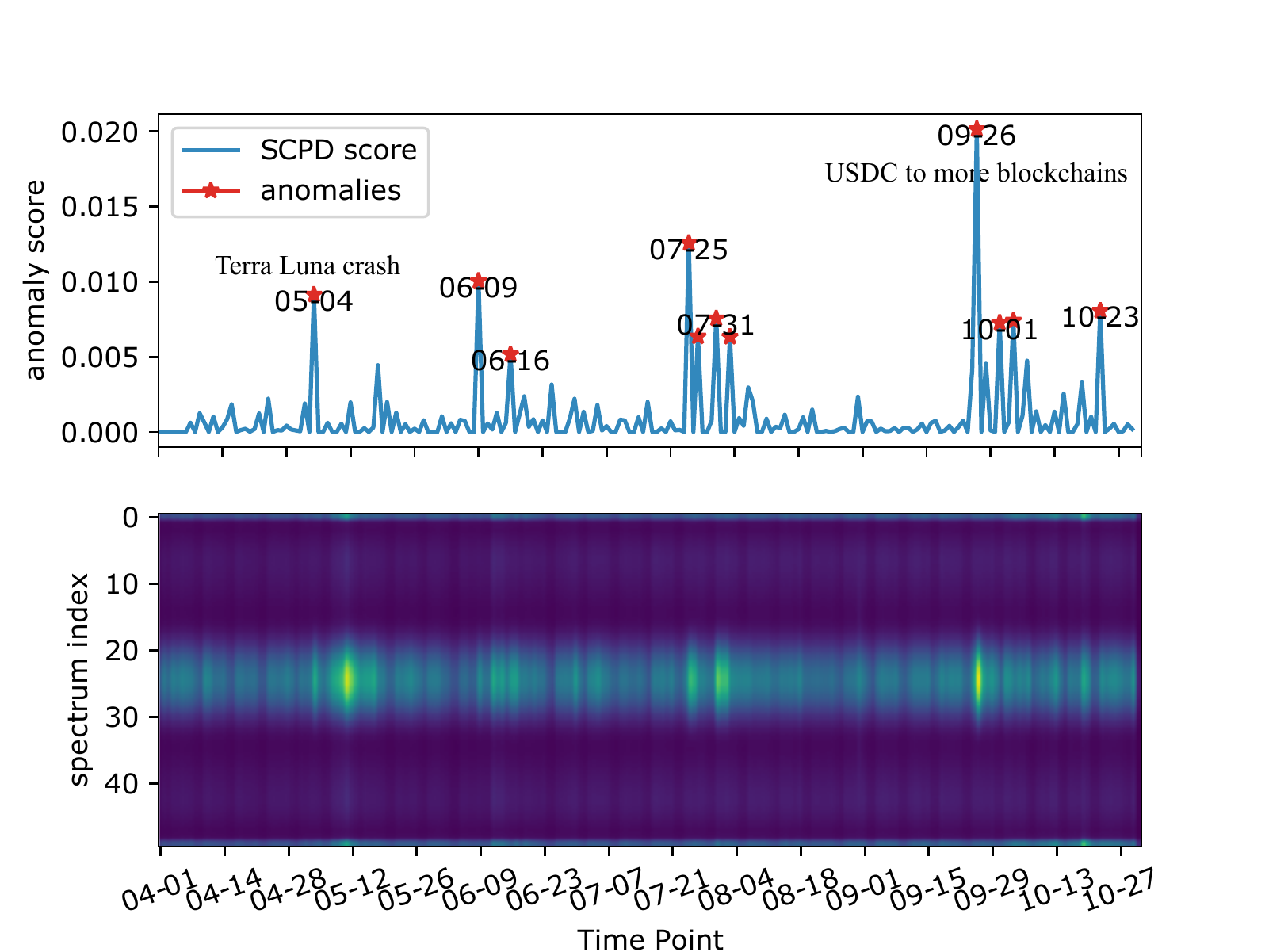}} 
        \caption{\newmethod detects the Lunar coin crash on the Stablecoin transaction datset.}
        \label{Fig:coin}
    \end{center}
    \vskip -0.4in
\end{figure*}

The Stablecoin transaction dataset~\footnote{\url{https://www.chartalist.org/eth/StablecoinAnalysis.html}}~\cite{shamsi2022chartalist} tracks the transaction network of six stablecoin networks. The nodes are contact addresses and the edges are transactions. The dataset spans from April first 2022 to November first 2022, and the most notable anomaly or change point is the Terra Luna crash when it lost its $\$1 $ USD fixed price value in early May 2022. Note that Stablecoins are special tokens that are meant to maintain a fixed price value, such as the $\$1 $ USD per token price mentioned above. Therefore, the Terra Luna crash has significant impact on the entire transaction network. Figure~\ref{Fig:coin} shows the detected change points by \newmethod. Most notably, \newmethod is able to detect two of the most significant events in the stablecoin network in the duration of the dataset. Notable, on May 4th, \newmethod shows the first significant jump in anomaly score, which corresponds to the Terra Luna Crash on May 7th. Signaling the anomaly 3 days early is a highly desirable behavior in transaction networks. In addition, the peak with the highest anomaly score occurs on September 26th when USDC coming to five additional blockchain ecosystems. More events are in Appendix~\ref{app:event}.

\section{Conclusion}

In this work, we proposed a novel change point detection method, \newmethod, to detect anomalous changes in the graph structure as well as node attributes in a dynamic graph. \newmethod approximates the distribution of Laplacian eigenvalues as an embedding for the graph structure and Local DOS embeddings to measure the alignment between node attributes and the eigenvectors of the Laplacian at different frequency intervals. In synthetic experiments, \newmethod achieves state-of-the-art performance while running efficiently on graphs with millions of edges. On three real-world datasets, \newmethod is able to capture structural and attribute change points corresponding to significant events. 

\begin{ack}

This research was supported by the Canadian Institute for Advanced Research (CIFAR AI chair program), Natural Sciences and Engineering Research Council of Canada (NSERC) Postgraduate Scholarship-Doctoral (PGS D) Award and Fonds de recherche du Québec - Nature et Technologies (FRQNT) Doctoral Award. We would like to thank Cuneyt G. Akcora and Kiarash Shamsi for their help with the Stablecoin Transaction Dataset experiment and discussing the significant events on this dataset.
\end{ack}

\bibliographystyle{abbrv}
\bibliography{110ref}

\newpage

\appendix

\section{Hyperparameter Sensitivity Study}
\label{app:sensitivity}

\begin{figure}[t]
\centering
\begin{subfigure}{0.5\columnwidth}
 \includegraphics[width=\columnwidth]{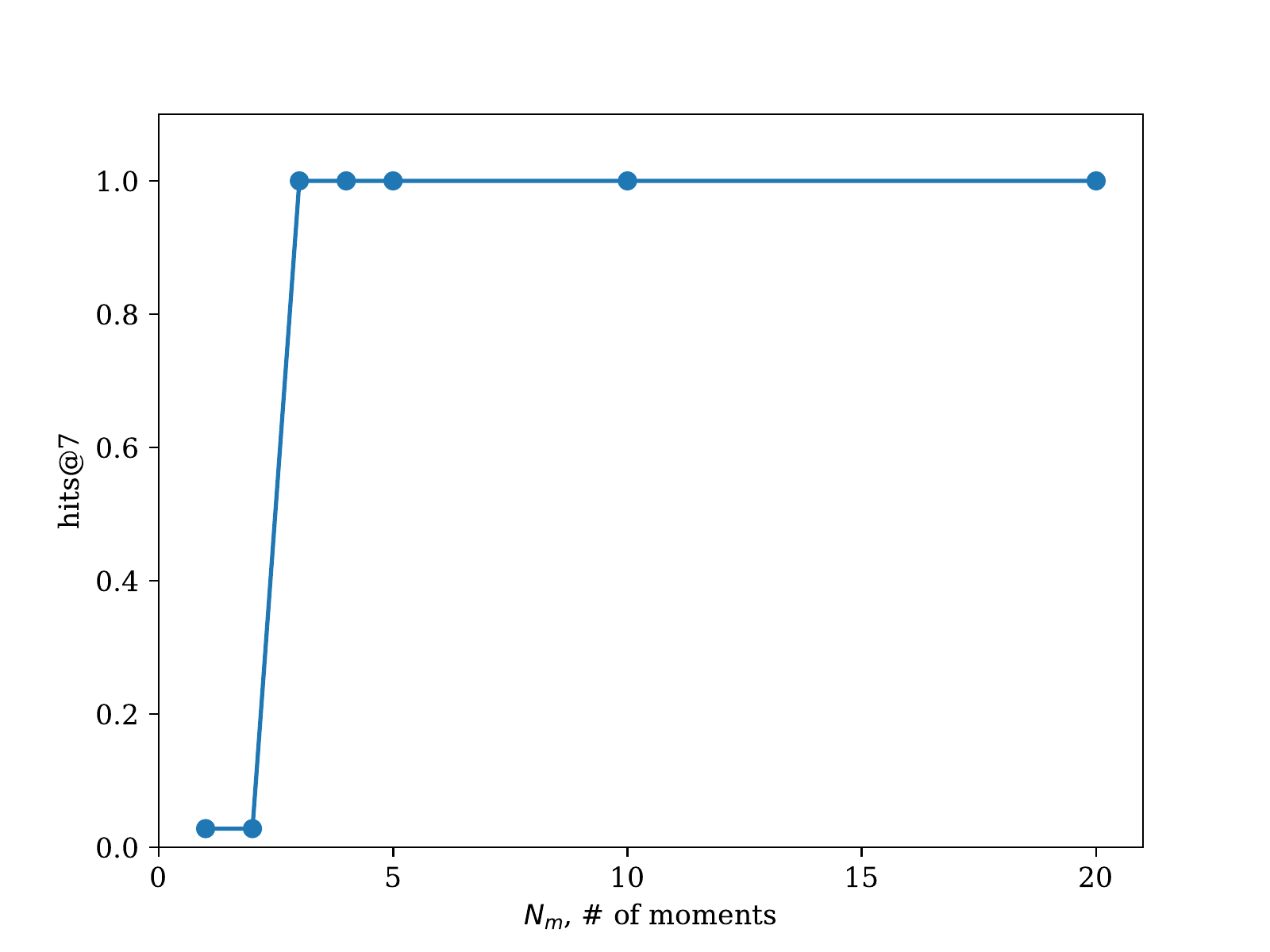}
  \caption{The number of Chebyshev moments $N_m$}
\end{subfigure}%
\begin{subfigure}{0.5\columnwidth}
  \centering
  \includegraphics[width=\columnwidth]{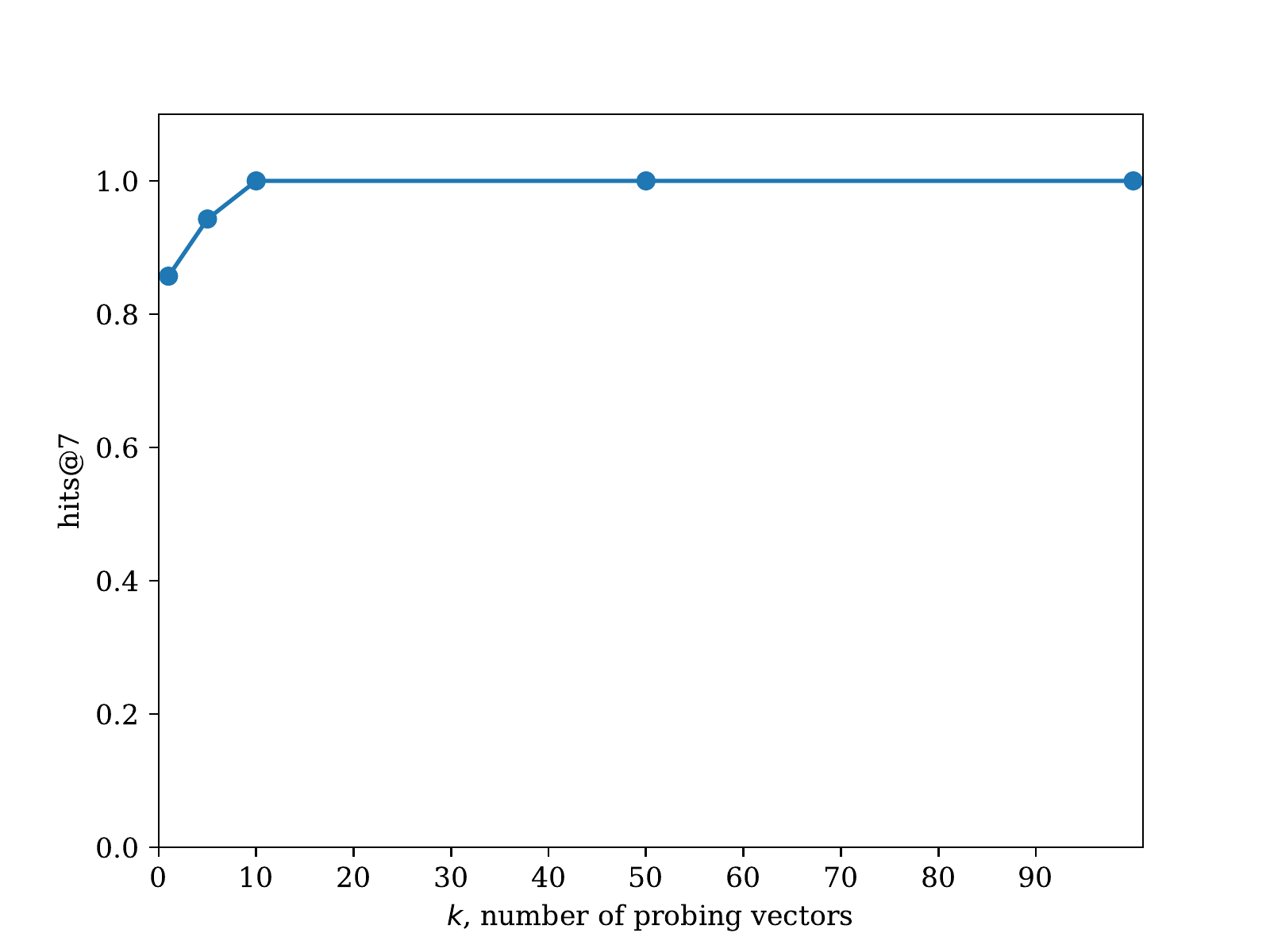}
  \caption{the number of probing vectors $N_z$}
\end{subfigure}%
\caption{\newmethod is robust to the change in hyperparameter including $N_m$ and $N_z$.}
 \label{Fig:sensitivity}
\vskip -0.2in
\end{figure}

In this section, we study the hyperparameter sensitivity of \newmethod. More specifically, we investigate the impact of the number of probing vectors $N_z$, the number of Chebychev moments $N_m$ and the number of equal-sized bins $k$ in the range of Laplacian eigenvalues on the performance of \newmethod. For all three hyperparameters, we see that \newmethod is robust to the choice of hyperparameters. In Figure~\ref{Fig:sensitivity}, we showed the Hits@7 performance of \newmethod in the SBM hybrid experiment averaged across 5 trials when $N_m$ and $N_z$ are varied~(details see Section~\ref{Sec:synthetic}. For the number of equal-sized bins $k$, \newmethod remains at 100\% Hits@7 performance when using a range of 5 to 50 bins. 

\section{Real World Experiment} \label{app:real}

We first provide more details about the real world datasets in Appendix~\ref{app:hist_data} and Appendix~\ref{app:flight_data}. Then, we provide additional results for \newmethod in Appendix~\ref{app:covid} and for baselines in Appendix~\ref{app:baseline_real}.

\subsection{MAG History Co-authorship Network} \label{app:hist_data}
MAG-History is a co-authorship dynamic network extracted from the Microsoft Academic Graph~(MAG)~\cite{Sinha-2015-MAG} by identifying publications which are marked with the "History" tag. This dataset was originally extracted in~\cite{Benson-2018-simplicial} as a temporal high-order network where nodes are authors and each simplex is the publication written by a group of authors. In this work, we project each simplex into a clique, thus forming an undirected dynamic graph. The dataset starts from 1837 to 2018 thus spanning 181 years. There are 2.8 million projected edges across all time steps and 0.7 million nodes in total.

\subsection{COVID Flight Network} \label{app:flight_data}
The COVID flight network~\footnote{\url{https://zenodo.org/record/3974209/\#.Yf62HepKguU}} is a derived dataset from the full OpenSky~\cite{schafer2014bringing, olive2019traffic} dataset to illustrate the development of air traffic during the COVID-19 pandemic. The nodes are airports and each edge is a timestamped tracked flight. We examine the period from 01-01-2020 to 07-27-2020 for a duration of 30 weeks. To reduce the noise and inconsistency due to the crowd-sourced nature of the dataset, we cleaned and processed the dataset further. First, we removed any flights without a known source or destination. Second, we only keep 3000 airports with the highest total degrees in the period of interest. Then, we use a full week as the duration of each snapshot to reduce the variability from daily flights. Lastly, we set the edges to be undirected and the weights to be how many days in a week this edge was observed, thus normalizing the edge weights to $[0,7]$. In this way, we are monitoring the existence and weekly frequency of flights between airports.


\subsection{\newmethod Results for Canadian Airports}\label{app:covid}
\begin{figure}[t]
    \begin{center}
        \centerline{\includegraphics[width=0.8\columnwidth]{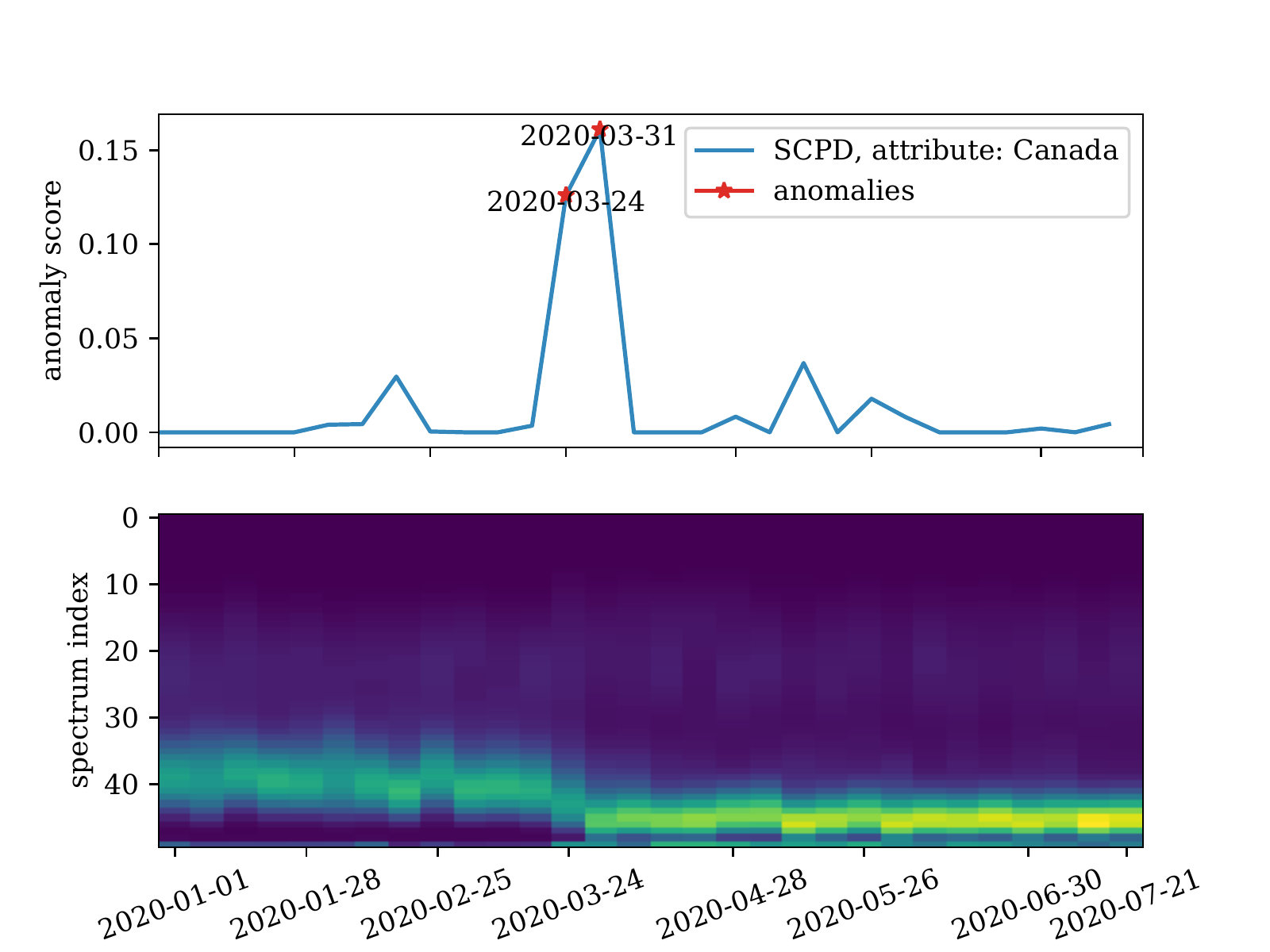}}
        \caption{\newmethod detects the effect of travel restrictions in Canada. On 03.18.2020, travel restrictions are applied to the entry of all foreign nationals (except US). On 03-21-2020, travel restrictions are implemented for the entry of all foreign nationals from US.}
        \label{Fig:covid.canada}
    \end{center}
    \vskip -0.3in
\end{figure}

We report additional results for the COVID Flight network. Here, we set the node attribute as an indicator vector for airports in Canada. In Figure~\ref{Fig:covid.canada}, \newmethod also correctly identifies the anomalies associated with Canadian airports. On 03-18, travel restrictions are applied to all foreign nationals~(except US) coming into Canada~\footnote{\href{https://www.cihi.ca/en/covid-19-intervention-timeline-in-canada}{intervention timeline in Canada}}. Then, on 03.21, travel restrictions are extended to foreign nationals coming from US as well. This figure also verifies that \newmethod can correctly detect specific change points to nodes with a given attribute.

\subsection{Additional Baseline Results} \label{app:baseline_real}

\begin{figure}[t]
    \begin{center}
        \centerline{\includegraphics[width=0.6\columnwidth]{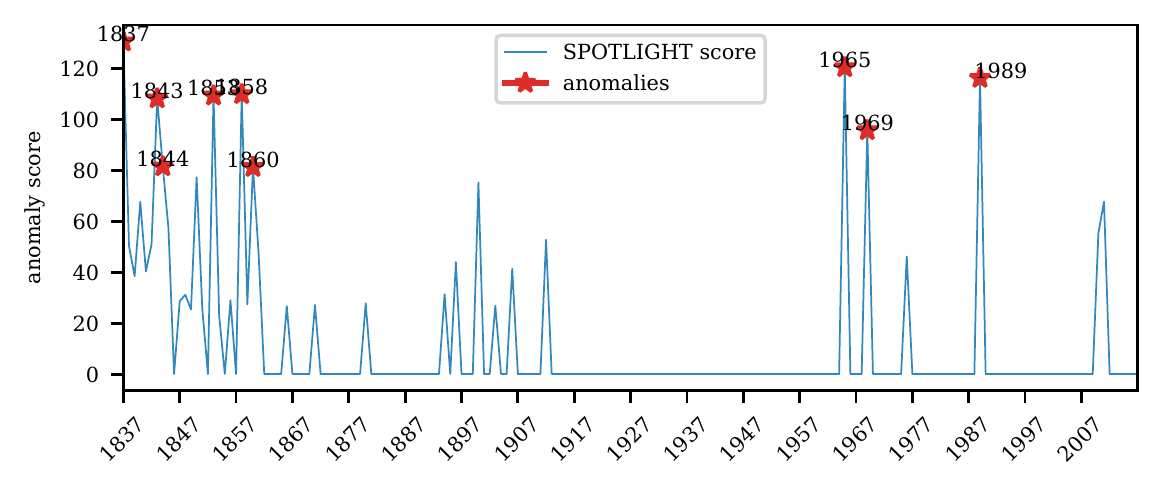}}
        \caption{SPOTLIGHT with rrcf result on the MAG History network. }
        \label{fig:sl_hist}
    \end{center}
\end{figure}

\begin{figure}[t]
    \begin{center}
        \centerline{\includegraphics[width=0.6\columnwidth]{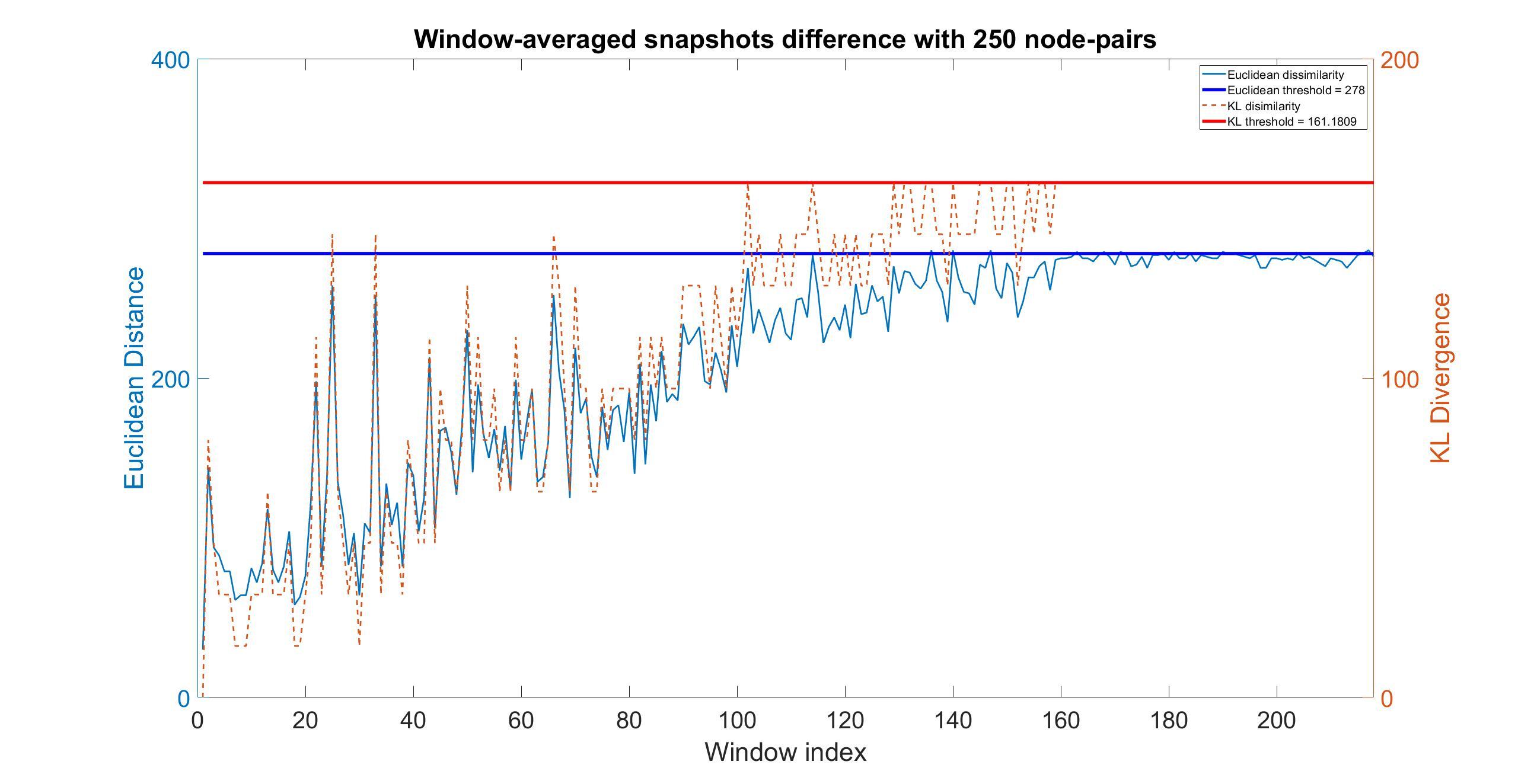}}
        \caption{EdgeMonitoring results on the MAG History dataset.}
        \label{fig:em_hist}
    \end{center}
\end{figure}

\begin{figure}[t]
    \begin{center}
        \centerline{\includegraphics[width=0.6\columnwidth]{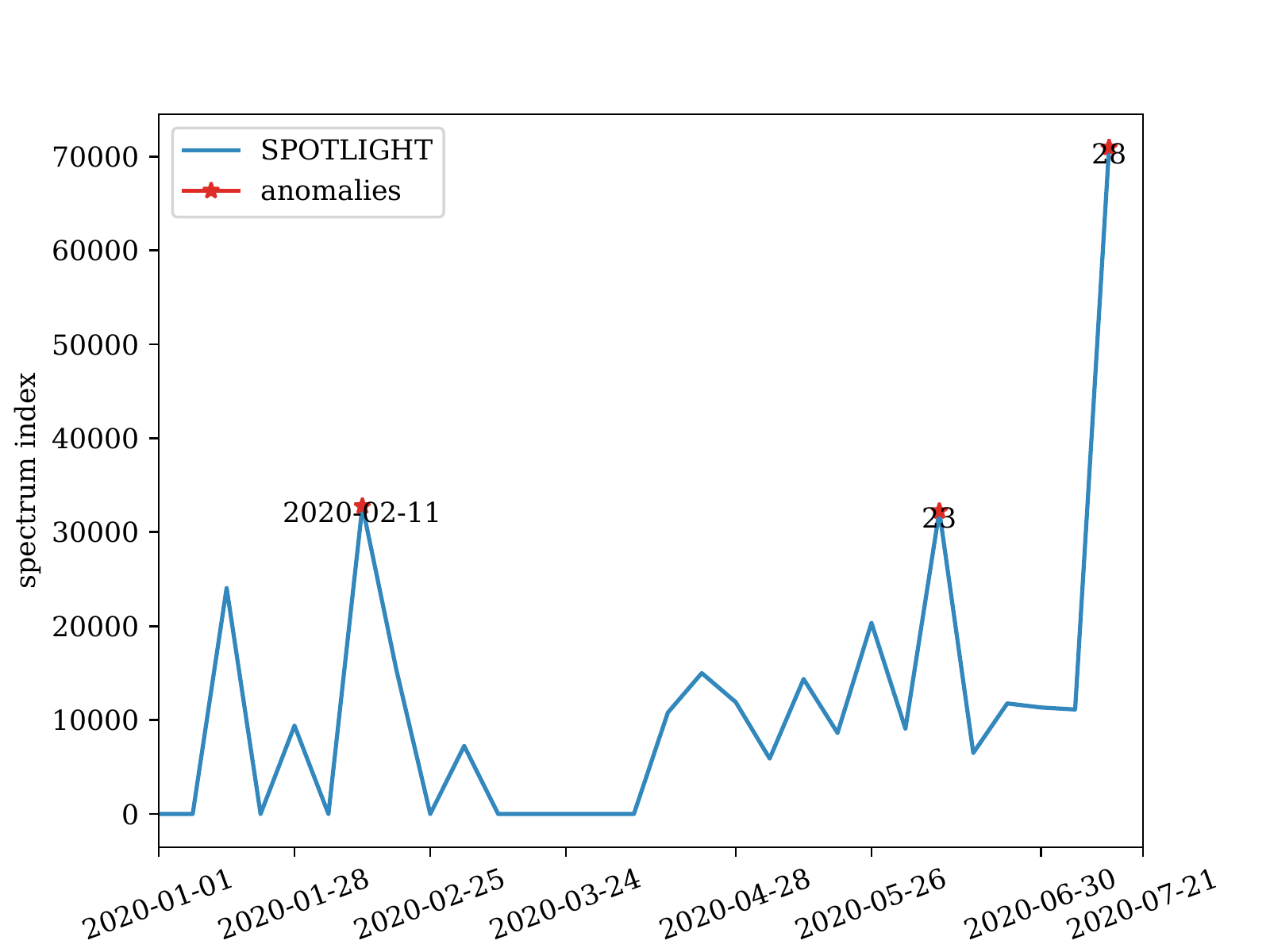}}
        \caption{SPOTLIGHT with sum result on the COVID flight network. }
        \label{fig:sl_flight}
    \end{center}
\end{figure}

\begin{figure}[t]
    \begin{center}
        \centerline{\includegraphics[width=0.6\columnwidth]{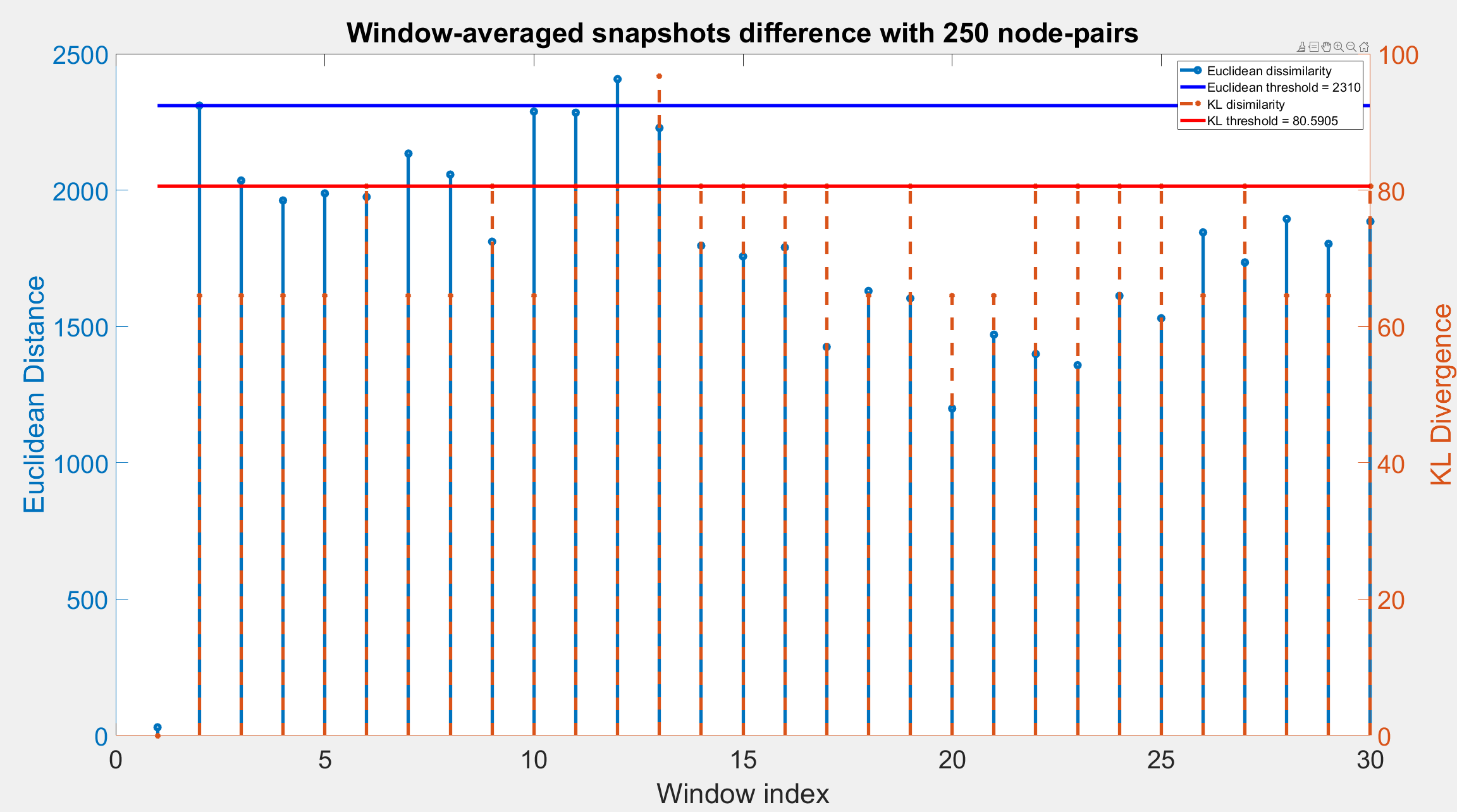}}
        \caption{EdgeMonitoring Results for the COVID flight network. }
        \label{fig:em_flight}
    \end{center}
\end{figure}

Here, we report the results of SPOTLIGHT and EdgeMonitoring for the MAG History Network~(see Figure~\ref{fig:sl_hist},~\ref{fig:em_hist}) and the COVID Flight Network~(see Figure~\ref{fig:sl_flight},~\ref{fig:em_flight}).

\subsection{Stablecoin Transaction Network Event List}\label{app:event}

\begin{table*}[t]
\centering 
\caption{ Full Event list in the Stablecoin Transaction Network as stated in \href{https://www.chartalist.org/eth/StablecoinAnalysis.html}{Chartalist}~\cite{shamsi2022chartalist}. The colored rows are the two most significant events from this list.}
\resizebox{\linewidth}{!}{%
  \begin{tabular}{l | l | l}
  \toprule
  Date & Type & Event \\ 
  \midrule
  \multicolumn{3}{c}{April 2022} \\
  4/11/2022 & positive & BlackRock and Fidelity Back USDC in 400 Million Funding Round \\
  4/19/2022 & positive & Terra UST takes over BUSD to become third largest stablecoin \\
  \midrule
  \multicolumn{3}{c}{May 2022} \\
  \rowcolor{orange!50}
  5/6/2022  & negative & LARGE amounts of UST selling on ANCHOR (approx 500million) \\
  5/7/2022  & negative & UST depegs LFG deploys assets to defend peg (750million in BTC sold to buy UST. BTC value drops) \\
  5/8/2022  & negative & UST Depegs again to 35 cents LUNA keeps falling \\
  5/11/2022 & negative & LUNA Collapses by almost 99 \% \\
  5/11/2022 & negative & UST Downfall Brings Record Volumes to Curve: IntoTheBlock \\
  5/12/2022 & negative & Terra Halts Its Blockchain and Binance Suspends LUNA Trading \\
  5/13/2022 & negative & LUNA essentially worthless (1/100th of a cent) \\
  5/15/2022 & negative & Luna Foundation Guard 80000 Bitcoin Lighter After Failing to Defend UST Peg \\
  5/26/2022 & positive & Tether Token (USDT) Launches on Polygon \\
  \midrule
  \multicolumn{3}{c}{June 2022} \\
  6/8/2022  & positive & Tether Tokens (USDT) to Launch on Tezos \\
  6/10/2022 & negative & Binance CEO mentions selling 500 million in FTX holdings\\
  6/12/2022 & positive & TRON DAO RESERVE DEPLOYS Funds (700million deployed) \\
  6/12/2022 & positive & USD 500 million in USDC added to TRON DAO USDD reserve \\
  \midrule
  \multicolumn{3}{c}{August 2022} \\
  8/7/2022 & negative & Circle freezes blacklisted Tornado Cash smart contract addresses \\
  8/10/2022 & unknown & FTX shows sign of trouble and Binance CEO and FTX come to non binding deal for a buyout of FTX \\
  8/10/2022 & negative & MakerDAO should seriously consider depegging DAI from USD — Founder \\
  8/21/2022 & negative & USDC whales leave and trading volume soars \\
  8/30/2022 & positive & Centre Consortium Announces Release of USD Coin Version 2.0 \\
  8/30/2022 & positive & MakerDAO Christensen Pushes Endgame Plan to Save DAI From Attack \\
  \midrule
  \multicolumn{3}{c}{September 2022} \\
  9/1/2022 & negative & Binance Announces LUNC and USTC Deposit And Withdrawal Suspension \\
  9/4/2022 & positive & Luna price has picked momentum as the community continues burning and staking LUNC tokens \\
  9/5/2022 & negative & FTX To Suspend LUNC and USTC Deposits And Withdrawals and Delist ANC \\
  9/6/2022 & negative & Binance - LUNC and USTC Deposits and Withdrawal Suspended \\
  9/8/2022 & positive & Huobi And BTCEX Joins Binance To Support 1.2\% Tax Burn For Terra Classic (LUNC) \\
  9/10/2022 & negative & Binance Backs out of Deal with FTX \\
  9/11/2022 & positive & Binance Announces Zero-Fee Trading For Luna \\
  9/11/2022 & positive & Tether (USDT) Launches on Near Network \\
  9/12/2022 & negative & FTX Suspends LUNA and UST Deposits and Withdrawals \\
  \rowcolor{orange!50}
  9/27/2022 & positive & USDC coming to five additional blockchain ecosystems \\
  \midrule
  \multicolumn{3}{c}{October 2022} \\
  10/6/2022 & positive & MakerDAO goes ahead with \$500M investment in treasuries and bonds \\
  10/8/2022 & positive & USTC jumps over 35\% in 24 hours as re-peg proposal emerges \\
  10/12/2022 & positive & Binance Adds UST to Borrowable Assets List \\
  10/12/2022 & positive & Tether change reserves to US treasury bills \\
  10/19/2022 & positive & UST Trading Vol Skyrockets By 240\% \\
  10/19/2022 & positive & Tether announces Available in Over 24000 ATMs Across Brazil on 3 November \\
  10/26/2022 & positive & UST climbs 15\% in 24 hours reclaiming \$ 400 million market cap \\
  \bottomrule
  \end{tabular}
}

\end{table*}

\end{document}